\pdfoutput=1
\documentclass[11pt]{article}

\usepackage[]{acl}

\usepackage{times}
\usepackage{latexsym}

\usepackage[T2A,T1]{fontenc}
\usepackage[utf8]{inputenc}
\usepackage[russian,english]{babel}

\usepackage{microtype}
\usepackage{tabularx}
\usepackage{multirow}

\usepackage{gb4e} \noautomath

\newcommand{\md}[1]{{\texttt{#1}}}

\usepackage{tikz}
\def\checkmark{\tikz\fill[scale=0.4](0,.35) -- (.25,0) -- (1,.7) -- (.25,.15) -- cycle;} 
\title{The SIGMORPHON 2022 Shared Task on Morpheme Segmentation}

\author{Khuyagbaatar Batsuren$^1$ \and G{\'a}bor Bella$^2$ \and Aryaman Arora$^3$ \and {\bf Viktor Martinovic}$^4$ \\ {\bf Kyle Gorman$^{5}$} \and {\bf Zdeněk Žabokrtský}$^6$ \and {\bf Amarsanaa Ganbold$^{1}$} \and {\bf Šárka Dohnalová}$^6$ \\ {\bf Magda Ševčíková}$^7$ \and {\bf Kateřina Pelegrinová}$^6$ \and {\bf Fausto Giunchiglia}$^2$ \and {\bf Ryan Cotterell}$^8$ \\ \and {\bf Ekaterina Vylomova}$^{9}$ \\ $^1$National University of Mongolia \hspace{.1cm} $^2$University of Trento \hspace{.1cm} $^3$Georgetown University \\ $^4$University of Vienna \hspace{.1cm} $^5$Graduate center, City University Of New York \hspace{.1cm} $^6$Charles University \\ $^7$University of Ostrava \hspace{.1cm} $^8$ETH Zürich \hspace{.1cm} $^9$University of Melbourne \\ 
}

\begin{document}
\maketitle
\begin{abstract}
The SIGMORPHON 2022 shared task on morpheme segmentation challenged systems to decompose a word into a sequence of morphemes and covered most types of morphology: compounds, derivations, and inflections. Subtask 1, word-level morpheme segmentation, covered 5 million words in 9 languages (Czech, English, Spanish, Hungarian, French, Italian, Russian, Latin, Mongolian) and received 13 system submissions from 7 teams and the best system averaged 97.29\% F1 score across all languages, ranging English (93.84\%) to Latin (99.38\%). Subtask 2, sentence-level morpheme segmentation, covered 18,735 sentences in 3 languages (Czech, English, Mongolian), received 10 system submissions from 3 teams, and the best systems outperformed all three state-of-the-art subword tokenization methods (BPE, ULM, Morfessor2) by 30.71\% absolute.
To facilitate error analysis and support any type of future studies, we released all system predictions, the evaluation script, and all gold standard datasets.\footnote{\url{https://github.com/sigmorphon/2022SegmentationST}}

\end{abstract}

\section{Introduction}
Many NLP applications, such as machine translation or question answering, require \emph{subword tokenization}, i.e. splitting words into a sequence of substrings \cite{mielke2021between}. Such tokenizers are trained by an unsupervised algorithm, usually either Byte-Pair Encoding (BPE; \citealt{gage1994new,sennrich2016neural}) or Unigram Language Modeling (ULM; \citealt{kudo-2018-subword}). To give a few examples, contemporary language models RoBERTa \cite{liu2019roberta} and GPT-3 \cite{brown2020language} use a byte-level BPE \cite{radford2019language} while XLNet \cite{yang2019xlnet} relies on ULM. These subword tokenization algorithms are not linguistically motivated but are rather based on statistical co-occurrences. Therefore, unsupervised and semi-supervised methods for morphological segmentation \cite{creutz2005unsupervised} have emerged in parallel, state-of-the-art methods of this kind being Morfessor variants \cite{gronroos2014morfessor,gronroos2020morfessor}. \citet{ataman2017linguistically} and \citet{schwartz2020neural} find that Morfessor-based language models can outperform BPE-based ones. \citet{matthews2018using,nzeyimana2022kinyabert} show that enriching BPE with morphological analyzers can be beneficial for translation, while many others \cite{domingo2018much,machavcek2018morphological,schwartz2020neural,saleva-lignos-2021-effectiveness} find no conclusive improvements over BPE for machine translation.

\begin{table}[t]
\small
\newcolumntype{R}{>{\raggedleft\arraybackslash}X}
\newcolumntype{C}{>{\centering\arraybackslash}X}
\newcolumntype{L}{>{\raggedright\arraybackslash}X}%%
\centering
\begin{tabularx}{\linewidth}{l|ll|r}

System & type   & motivation   & segmentation             \\
\hline
\hline
BPE  & surface & sta. & in | val | uable       \\
Morfessor2  & surface    & sta. \& lin. & in | valuable         \\
DeepSPIN-3  & canonical & sta. \& lin. & in | value | able      \\

\hline
\end{tabularx}
\caption{\label{tab:examples} Structural differences of subword tokenization (BPE), morphological segmentation (Morfessor2), and morpheme segmentation (DeepSPIN-3 -- subtask 1 winning system); acronyms: sta. - statistics and lin. - linguistic}
\end{table}

One of the core problems is that the state-of-the-art morphological segmentation and subword tokenization algorithms provide  ``surface-level'' segmentation, which has several theoretical drawbacks  with respect to ``canonical'' segmentation (e.g., segmented substrings are not considered as meaningful as morphemes). \citet{cotterell2016joint} provided formal definitions for both:
given a word $w$, its ``surface'' segmentation is a sequence of \textit{surface substrings} the concatenation of which is~$w$, e.g., \textit{funniest} → \textit{funn-i-est}. The purpose of canonical segmentation \cite{kann-etal-2016-neural,Task2-TueSeg}, on the other hand, is not only computing surface segmentation but also restoring standardized forms of morphemes, e.g., \textit{funniest} → \textit{fun-y-est}. More detailed structural distinctions between these segmentation types are shown in Table~\ref{tab:examples}. 

However, state-of-the-art studies in canonical segmentations have been limited to very low numbers of languages with sufficiently rich morphological resources \cite{kurimo2010morpho,kurimo2010proceedings,cotterell2016joint,kann-etal-2018-fortification}. With the goal of advancing research in this direction, we present a \textit{morpheme segmentation shared task} and provide large-scale datasets over nine languages, evaluation metrics, and morphological annotations of five million word formations. In this, we rely on the latest release of UniMorph \cite{batsuren2022unimorph} which has introduced morpheme segmentations and derivational data from MorphyNet \cite{batsuren-etal-2021-morphynet}. The resulting shared task is a follow-up to past morphological segmentation shared tasks such as ``MorphoChallenge'' \cite{kurimo2007unsupervised,kurimo2008overview,kurimo2009overview} or ``Multilingual parsing''  \cite[where lemmatization as segmentation is a subtask]{zeman2017conll}.

\begin{table}[t]
\small
\newcolumntype{R}{>{\raggedleft\arraybackslash}X}
\newcolumntype{C}{>{\centering\arraybackslash}X}
\newcolumntype{L}{>{\raggedright\arraybackslash}X}%%
\centering
\begin{tabularx}{\linewidth}{clLc}
\hline
Lang & word      & segmentation             & category \\
\hline
\multirow{2}{*}{eng}  & sheepiness & sheep @@y @@ness   & 010      \\
  & pokers     & poke @@er @@s      & 110      \\
\hline
\multirow{2}{*}{hun}  & időpontod  & idő @@pont @@od    & 101      \\
  & szőttetek  & sző @@tt @@etek    & 100      \\
\hline
\multirow{2}{*}{mon}  & \foreignlanguage{russian}{харах}      & \foreignlanguage{russian}{харах}              & 000     \\
  & \foreignlanguage{russian}{гэмтлийг}   & \foreignlanguage{russian}{гэмтэх @@л @@ийг} & 110      \\

\hline
\end{tabularx}
\caption{\label{tab:training_examples} Training samples for Subtask 1. Each sample consists of a word, its canonical segmentation, and a category encoding word formation processes.}
\end{table}

\begin{table*}[h]
\small
\newcolumntype{R}{>{\raggedleft\arraybackslash}X}
\newcolumntype{C}{>{\centering\arraybackslash}X}
\newcolumntype{L}{>{\raggedright\arraybackslash}X}%%
\centering
\begin{tabularx}{\textwidth}{|c|ccc|l|L|}
\hline
Category & Infl. & Deri. & Comp. & Description  & English example (input ==\textgreater ~ output)     \\
\hline
000        & - & - & - &Root words (free morphemes)                     & progress ==\textgreater ~progress                  \\
100        & \checkmark & - & - & Inflection only                  & prepared ==\textgreater  ~ prepare @@ed                   \\
010        & - & \checkmark & - &Derivation only                  & intensive ==\textgreater ~intense @@ive                   \\
001        & - & - & \checkmark &Compound only                    & hotpot ==\textgreater ~hot @@pot                   \\
101        & \checkmark & - & \checkmark & Inflection and Compound          & wheelbands ==\textgreater ~wheel @@band @@s        \\
011        & - & \checkmark & \checkmark & Derivation and Compound          & tankbuster ==\textgreater ~tank @@bust @@er        \\
110        & \checkmark & \checkmark & - & Inflection and Derivation        & urbanizes ==\textgreater ~urban @@ize @@s          \\
111        & \checkmark & \checkmark & \checkmark & Inflection, Derivation, Compound & trackworkers ==\textgreater ~track @@work @@er @@s\\
\hline
\end{tabularx}
\caption{\label{tab:categories}Morphological categories and descriptions of segmented words in subtask 1}
\end{table*}

\begin{table*}[ht]
\small
\newcolumntype{R}{>{\raggedleft\arraybackslash}X}
\newcolumntype{C}{>{\centering\arraybackslash}X}
\newcolumntype{L}{>{\raggedright\arraybackslash}X}%%
\centering
\begin{tabularx}{0.95\textwidth}{|C|rrrrrrrrr|}
\hline
Category & English & Spanish & Hungarian & French & Italian & Russian & Czech & Latin  & Mongolian \\
\hline
000         & 101938  & 15843   & 6952      & 13619  & 21037   & 2921    & -     & 50338  & 1604      \\
100         & 126544  & 502229  & 410662    & 105192 & 253455  & 221760  & -     & 831991 & 7266      \\
010         & 203102  & 18449   & 24923     & 67983  & 41092   & 72970   & -     & 0      & 2201      \\
001         & 16990   & 248     & 3320      & 1684   & 431     & 259     & -     & 0      & 5         \\
101         & 13790   & 458     & 101189    & 478    & 317     & 1909    & -     & 0      & 35        \\
011         & 5381    & 82      & 1654      & 506    & 140     & 328     & -     & 0      & 0         \\
110         & 106570  & 346862  & 323119    & 126196 & 237104  & 481409  & -     & 0      & 7855      \\
111         & 3059    & 343     & 54279     & 186    & 158     & 2658    & -     & 0      & 0         \\
\hline
total words & 577374  & 884514  & 926098    & 382797 & 553734  & 784214  & 38682 & 882329 & 18966    \\
\hline
\end{tabularx}
\caption{\label{tab:task1stats}Word statistics across morphological categories on subtask 1} 
\end{table*}

\section{Task and Evaluation Details}
\subsection{Subtask 1: Word-level Morpheme Segmentation}
In subtask 1, participating systems were asked to segment a given word into a sequence of morphemes. The participants were initially provided with examples of segmentation to train and fine-tune their systems, as shown in Table~\ref{tab:training_examples}. Each instance in the training set is a triplet consisting of a word, a sequence of morphemes, and a morphological category specifying the types of word formation (see Table~\ref{tab:categories}). The morphological category is an optional feature that can only be used to oversample or undersample the training dataset (the word frequencies are imbalanced across the morphological categories, e.g., Italian has 431 compound words and 253K inflections). The test data only contained the initial word itself.
%Table~ \ref{tab:categories} shows more detailed description of this morphological categories. 
%At the release time, we provided only first column of the test datasets as input words, and all three columns of the development datasets. 

Key points of this subtask are:
\begin{itemize}
\item The task is focusing on canonical segmentation, i.e. given an input word, participants had to predict \emph{a sequence of morphemes}. In canonical segmentation, the participating systems need to reconstruct internal morphophonological processes involved in word formation. For example, the word ``intensive'' will be decomposed into the base form ``intens\textit{\textbf{e}}''  and the adjectival siffix `@@ive'' (note that the ending `\textit{\textbf{e}}' of the base word is inferred here);
\item As shown in Table~\ref{tab:task1stats}, the task is multilingual, with seven high-resource languages (English, Spanish, Hungarian, French, Italian, Russian, Latin) and two low-resource languages (Czech and Mongolian);
\item The annotated corpus data represents a variety of morphological phenomena, including inflection, derivation, compounding (Table \ref{tab:task1stats});
\item A large-scale coverage as segmentations of five million words.
\end{itemize}

%In each language's data of this subtask, we tried to cover all types of morphologically complex words including root words, compound words, derivations, and inflected forms.  

\subsection{Subtask 2: Sentence-level Morpheme Segmentation}
The second subtask is a context-dependent morpheme segmentation and focuses on resolving ambiguity in segmentations. Consider the following example containing a Mongolian homonym:
%A successful system would predict sequences of morphemes from ambiguous words in a given sentence. The main reason is that morpheme segmentation of words in some cases is entirely dependent on given contexts. Let us consider the following example of a Mongolian homonym.
\begin{exe}
\ex
\glll \foreignlanguage{russian}{Гэрт} \foreignlanguage{russian}{эмээ} \foreignlanguage{russian}{хоол} \foreignlanguage{russian}{хийв}\\
\foreignlanguage{russian}{Гэр @@т} \foreignlanguage{russian}{эмээ} \foreignlanguage{russian}{хоол} \foreignlanguage{russian}{хийх @@в} \\
Home.\texttt{DAT} grandma meal cook.\texttt{PRS.PRF} \\
\glt `Grandma just cooked a meal at home.'
\end{exe}

\begin{exe}
\ex
\glll \foreignlanguage{russian}{Би} \foreignlanguage{russian}{өдөр} \foreignlanguage{russian}{эмээ} \foreignlanguage{russian}{уусан }\\
\foreignlanguage{russian}{Би} \foreignlanguage{russian}{өдөр} \foreignlanguage{russian}{эм @@ээ} \foreignlanguage{russian}{уух @@сан} \\
I afternoon medicine.\texttt{PSSD} take.\texttt{PST} \\
\glt `Afternoon I took my medicine.'
\end{exe}

\noindent where ``\foreignlanguage{russian}{эмээ}'' is a homonym of two different words; in the first sentence, it is ``grandmother'', and in the second sentence --- an inflected form of ``medicine''. Thus, the form in the second case can be segmented. However, the modern subword segmentation tools consider no contextual differences in word forms.
%Maybe because of this reason, Google Translate fails to translate the second sentence as the translation output\footnote{accessed at translate.google.com on 03/June/2022.} is `I drank my grandma that day.'  

Key points of this subtask are:
\begin{itemize}
\item Morpheme segmentation is context-dependent;
\item We organize it for three languages: English, Czech, and Mongolian;
\item For Czech and Mongolian we asked native speakers to manually annotate the data. The details of data collection are provided in Section~\ref{sec:Data}.

\end{itemize}

\begin{table}[t]
\newcolumntype{R}{>{\raggedleft\arraybackslash}X}
\newcolumntype{C}{>{\centering\arraybackslash}X}
\newcolumntype{L}{>{\raggedright\arraybackslash}X}%%
\centering
\begin{tabularx}{0.9\linewidth}{lRRR}
\hline
Language & train & dev  & test \\
\hline
Czech     & 1,000  & 500  & 500  \\
English   & 11,007 & 1,783 & 1,845 \\
Mongolian & 1,000  & 500  & 600 \\

\hline
\end{tabularx}
\caption{\label{tab:task2stats}The number of samples in each language in Subtask 2.}
\end{table}

\subsection{Evaluation}
In order to evaluate and compare the systems, we used four metrics: (i) \textit{\textbf{precision}}, the ratio of correctly predicted morphemes over all predicted morphemes; (ii) \textit{\textbf{recall}}, the ratio of correctly predicted morphemes over all gold-label morphemes; (iii) \textit{\textbf{f-measure}},  the harmonic mean of the precision and recall;
(iv) \textit{\textbf{edit distance}} - average Levenshtein distance between the predicted output and the gold instance. For convenience, we provided the python tool\footnote{\url{https://github.com/sigmorphon/2022SegmentationST/tree/main/evaluation}} to evaluate these metrics on both subtasks. In addition, for subtask 1 this tool also provided detailed results across the morphological categories.

\section{Data}
\label{sec:Data}
We collected our morphological data from various sources to account for all types of morphology: derivational, inflectional, compounding. We also collected base forms. For derivational and inflectional morphology, we have used the segmentation data from UniMorph 4.0 \cite{batsuren2022unimorph} and MorphyNet \cite{batsuren-etal-2021-morphynet}. UniMorph contains inflectional paradigms collected from linguistic sources as well as Wiktionary, while MorphyNet represents derivations scraped from various editions of Wiktionary. Compounds and base forms were also extracted from Wiktionary (see  Section~\ref{sub:extraction} for more details on the data extraction). We then used the data to produce morpheme segmentations for seven high-resource languages. For Czech and Mongolian, as low-resource languages, we asked native speakers and linguists to develop the resources (Section~\ref{sub:LRL} provides more details). For English sentence data, we have used the universal dependency treebank of English \cite{silveira14gold}. 

\subsection{Data Statistics}
The data for the shared task was moderately multilingual, containing nine unique languages of five genera including Germanic, Italic, Slavic, Mongolic, and Uralic. In subtask 1, we have over 5 million samples of morpheme segmentations that cover nine languages over nine morphological categories, as shown in Table~ \ref{tab:task1stats}. In  subtask 2, Table~\ref{tab:task2stats} displays the data statistics of three languages. 

\subsection{Extraction from Wiktionary}
\label{sub:extraction}
Language-specific editions of Wiktionary contain a considerably large amount of derivations and compounds. 

\emph{Compound extraction rules} were applied to the etymology sections of Wiktionary entries to collect the Morphology template usages, such as for the English \emph{newspaper}:
\begin{center}
     Equivalent to \textbf{news} + \textbf{paper}.
\end{center}
where we have a morphology entry from the Wiktionary XML dump as follows:
\begin{center}
\{\{compound~|~en~|~news~|~paper\}\} 
\end{center}
Most of compound entries use ``compound'' etymology template while some cases use ``affix`` templates, e.g., \emph{basketball} and \emph{volleyball}. 

\emph{Root (and base) word extraction} is a two-step procedure. In the first step we collected words, inherited from earlier phases of corresponding languages. For example, English `book' is traced back to the Middle English `bok', according to the etymology section of Wiktionary. We extracted 279,173 words from 6 languages from CogNet, a cognate database containing 8.1 million cognate pairs of 335 languages from Wiktionary  \cite{batsuren2019cognet,batsuren2021large}. In the second step, we filtered out 116,863 words from the earlier extracted derivational and compound data, resulting in 162,310 root words in 6 languages. Similar Wiktionary data extraction procedures have been applied to a wide range of linguistic data, e.g., etymology  \cite{fourrier2020methodological}, multilingual lexicons - DBnary \cite{serasset2015dbnary} and Yawipa \citep{wu-yarowsky-2020-computational}.  
\begin{table*}[t]
\small
\newcolumntype{R}{>{\raggedleft\arraybackslash}X}
\newcolumntype{C}{>{\centering\arraybackslash}X}
\newcolumntype{L}{>{\raggedright\arraybackslash}X}%%
\centering
\begin{tabularx}{\textwidth}{l|l|l|CcCcc}
     &  &      & \multicolumn{5}{c}{System features} \\
Team     & Description & System     & Neural & Ensemble & Data+ & Multilingual & Multi-task \\
\hline
\hline
\multirow{3}{*}{Baseline} & \cite{schuster2012japanese}           & \md{WordPiece*}        & -      & -        & -     & -            & -          \\
 & \cite{kudo-2018-subword}           & \md{ULM*}        & -      & -        & -     & -            & -          \\
 & \cite{virpioja2013morfessor}           & \md{Morfessor2*}        & -      & -        & -     & -            & -          \\
\hline
\hline
\multirow{6}{*}{AUUH}     &    \multirow{6}{*}{\cite{auuh22sigmorphon}}       & \md{AUUH\_A*}    & \checkmark    &    -      & \checkmark   & \checkmark          & \checkmark        \\
         &             & \md{AUUH\_B*}    & \checkmark    &     -     & -   & \checkmark          & \checkmark        \\
         &             & \md{AUUH\_C}    & \checkmark    &       -   & \checkmark   &          -    & \checkmark        \\
         &             & \md{AUUH\_D}    & \checkmark    &      -    & -   &        -      & \checkmark        \\
         &             & \md{AUUH\_E*}    & \checkmark    &       -   & \checkmark   &        -      &     -       \\
         &             & \md{AUUH\_F*}    & \checkmark    &     -     & -   &        -      &     -       \\
\hline
\hline
\multirow{4}{*}{CLUZH}    &   \multirow{4}{*}{\cite{cluzh_sig22}}          & \md{CLUZH}      & \checkmark    & \checkmark      &    -   &        -      &     -       \\
         &             & \md{CLUZH-1}    & \checkmark    & \checkmark      &   -    &        -      &    -        \\
         &             & \md{CLUZH-2}    & \checkmark    & \checkmark      &     -  &         -     &    -        \\
         &             & \md{CLUZH-3}    & \checkmark    & \checkmark      &    -   &          -    &   -         \\
\hline
\hline
\multirow{3}{*}{DeepSPIN} &  \multirow{3}{*}{\cite{DeepSPIN2022}}           & \md{DeepSPIN-1} & \checkmark    &     -     &  -     &      -        &      -      \\
         &             & \md{DeepSPIN-2} & \checkmark    &    -     &     -  &     -         &   -         \\
         &             & \md{DeepSPIN-3} & \checkmark    &   -       &  -     &       -       &      -      \\
\hline
\hline
\multirow{2}{*}{GU}      &  \multirow{2}{*}{\cite{GU2022}}           & \md{GU-1}       & \checkmark    &     -     & \checkmark   &        -      &     -       \\
         &             & \md{GU-2}       & \checkmark    &      -    & \checkmark   &      -        &   -         \\
\hline
\hline
NUM DI   &  \cite{Task2_NUMDI}           & \md{NUM DI}     & \checkmark    &     -     &     -  &           -   &    -        \\
\hline
\hline
JB132    &  \cite{JB132}          & \md{JB132}      &    -   &       -   &     -  &      -        &         -   \\
\hline
\hline
\multirow{2}{*}{Tü Seg}  &  \multirow{2}{*}{\cite{Task2-TueSeg}}        & \md{Tü\_Seg-1}    & \checkmark    &    -      &     -  &       -     & -

\\ 
&           &  \md{Tü\_Seg-2} & \checkmark    &    -      &     -  &         -     & \checkmark  
\end{tabularx}
\caption{\label{tab:systems}The list of participating systems submitted to the shared task and baseline systems; Systems marked with * are submitted to both subtasks}
\end{table*}

\subsection{Collecting data for Czech and Mongolian}
\label{sub:LRL}
We had two languages with limited amount of data, Czech and Mongolian. For each  language, we used a different development methodology than for the other seven languages
(with larger amount of available data).

\textbf{Mongolian}: we asked two linguists (who are also native speakers of Mongolian) to annotate morpheme segmentations of 3,810 words from Mongolian WordNet \cite{batsuren-etal-2019-building}. After manual annotation, we received 1,604 base forms, 2201 derived forms, and 5 compounds. To account for inflectional morphology, we have used the Mongolian transducer tool \cite{munkhjargal2016morphological} to generate inflected forms of the 3,810 annotated words. In total, we collected morpheme segmentations of 18,966~Mongolian words for subtask~1. For subtask~2, the same two linguists annotated 2,100~Mongolian sentences.

\textbf{Czech}: we merged hand-segmented word forms from four sources for the purpose of subtask 1:
(a) segmentations previously created within DeriNet \cite{derinet-2019}, a project aimed at capturing derivational relations in Czech  (9,508 word forms),
(b) segmentations of Czech verb lemmas imported from a partially digitized version of a printed dictionary (\citealt{slavickova-2017}; 13,162 word forms in addition, i.e. not counting overlaps),
(c) segmentations available in the MorfCzech dataset  \cite{morfoczech-data-2022}, mostly extracted from dictionaries and  grammar books existing for Czech (additional 11,137 word forms), and
(d) word forms that we annotated newly in order to reach complete coverage of Czech subtask 2 sentences (see below; additional 4,887 word forms). In total, the subtask~1 dataset contains 38,694 unique Czech word forms segmented to morphs.

All annotations were performed by native speakers with linguistic education, and underwent careful harmonization if the input resources disagreed, as well as numerous consistency checks. However, because of rich allomorphy in Czech, we have not been able to merge allomorph sets  under more abstract umbrella morphemes so far, and thus words are represented as sequences of morphs (whose concatenation perfectly matches the original word forms), not of morphemes. 

The Czech subtask~2 dataset contains in total 2,000 sentences from the Czech subset of Universal Dependencies (\citealt{ud-cl-2021}; more specifically, 1000, 500, and 500 first sentences from the train, dev, and test sections, respectively, of the Prague Dependency Treebank subset of UD 2.9). Given that homonymy resulting in different morph boundaries is extremely rare in Czech, words are segmented basically regardless of their contexts.

\subsection{Data Splits}
From each language's collection of morpheme segmentations in subtask 1, we sampled 80\% for the training,  10\% for development, and 10\% for test sets.\footnote{All the data splits can be obtained from \url{ https://github.com/sigmorphon/2022SegmentationST/tree/main/data}} All splits of subtask 1 are balanced w.r.t. the nine morphological categories, described in Table~ \ref{tab:categories}. While sampling the training and development sets for the subtask 1, we excluded words that were present in the test sentences of subtask 2. This was done in order to avoid situations when the subtask 1 data could directly influence the results of subtask 2 (since we allowed the multi-task learnings between both subtasks).  

\section{Baseline Systems}
The shared task provided predictions and results of baseline systems to participants that covered all languages and both subtasks. We chose three baseline systems: 
First is \texttt{WordPiece},  one of the state-of-the-art subword tokenization algorithms used in BERT \cite{devlin-etal-2019-bert}, which is based on \citet{schuster2012japanese} and somewhat resembles BPE \cite{sennrich2016neural}. Second is \texttt{ULM} (Unigram Language Model \citet{kudo-2018-subword}), another popular subword tokenization, used in XLNet \cite{yang2019xlnet}. Third is \texttt{Morfessor2}, one of the state-of-the-art unsupervised morphological segmentations \cite{virpioja2013morfessor}. 

In future shared tasks, we aim to include more state-of-the-art tokenization tools including other Morfessor variants \cite{gronroos2014morfessor,ataman2017linguistically,gronroos2020morfessor}, BPE-dropout \cite{provilkov2019bpe}, dynamic programming encoding (DPE) \cite{he2020dynamic} or its variant \cite{hiraoka2021joint,song2022self}, multi-view subword regularization \cite{wang2021multi},  Charformer \cite{tay2021charformer}, space-treatment variants of BPE and ULM \cite{gow2022improving}.
\begin{table*}[t]
\small
\newcolumntype{R}{>{\raggedleft\arraybackslash}X}
\newcolumntype{C}{>{\centering\arraybackslash}X}
\newcolumntype{L}{>{\raggedright\arraybackslash}X}%%
\centering
\begin{tabularx}{\textwidth}{l|RRRRRRRRR|R}
     &             &             &             &             &             &             &             &             &             & macro      \\
System     & \multicolumn{1}{c}{ces}            & \multicolumn{1}{c}{eng}            & \multicolumn{1}{c}{fra}            & \multicolumn{1}{c}{ita}            & \multicolumn{1}{c}{lat}            & \multicolumn{1}{c}{rus}            & \multicolumn{1}{c}{mon}            & \multicolumn{1}{c}{hun}            & \multicolumn{1}{c|}{spa}            & avg.     \\
\hline
\hline
WordPiece       & 20.42          & 23.06          & 12.66          & 9.08           & 8.84           & 13.81          & 14.58          & 24.00          & 16.57          & 15.89          \\
ULM      & 23.71 & 32.32 & 16.08 & 10.65 & 10.42 & 15.67 & 25.82 & 31.27 & 19.58 & 20.61
          \\
Morfessor2 & 29.43 & 37.65 & 22.38 & 9.02 & 14.53 & 17.71 & 37.80 & 40.96 & 20.64 & 25.57\\

\hline
\hline
AUUH\_A*    & 93.65          & 92.32          & -              & -              & -              & -              & 98.19          & -              & -              & 94.72          \\
AUUH\_B*    & 93.85          & 93.20          & -              & -              & -              & -              & 98.31          & -              & -              & 95.12          \\
AUUH\_E*   & 90.71          & 87.10          & 90.78          & 92.39          & 98.71          & 94.33          & 96.06          & -              & -              & 92.87          \\
AUUH\_F    & 90.28          & 86.40          & 90.81          & 92.56          & 98.85          & 93.68          & 95.32          & 98.34          & 97.25          & \textbf{93.72} \\
\hline
\hline
CLUZH   & 93.81          & 92.70          & 94.80          & 96.93          & 99.37          & 98.62          & 98.12          & 98.54          & 98.74          & \textbf{96.85} \\
\hline
\hline
DeepSPIN-1 & 93.42          & 92.29          & 91.66          & 96.01          & 99.37          & 98.75          & 98.03          & 98.56          & 98.79          & 96.32          \\
DeepSPIN-2 & \textbf{93.88} & 93.39          & 95.29          & \textbf{97.47} & 99.36          & 99.30          & 98.00          & 98.68          & 99.02          & 97.15          \\
DeepSPIN-3 & 93.84          & \textbf{93.63} & \textbf{95.73} & 97.43          & \textbf{99.38} & \textbf{99.35} & \textbf{98.51} & \textbf{98.72} & \textbf{99.04} & \textbf{97.29} \\
\hline
\hline
GU-1*      & -              & -              & 83.44          & 88.69          & -              & -              & -              & -              & -              & 86.07          \\
GU-2*     & -              & -              & 83.38          & 87.49          & -              & -              & -              & -              & 95.95          & 88.94          \\
\hline
\hline
JB132      & 64.65          & 65.43          & 46.20          & 33.44          & 91.39          & 50.55          & 57.82          & 72.64          & 43.39          & \textbf{58.39} \\
\hline
\hline
NUM DI*     & -          & 83.56          & -              & 89.55          & -              & -              & 85.59          & 95.91          & -              & 88.65         \\
\hline
\hline
Tü\_Seg-1    & 93.38          & 90.51          & 93.76          & 95.73          & 99.37          & 98.21          & 97.02          & 98.59          & 97.93          & \textbf{96.06}
\end{tabularx}
\caption{\label{tab:subtask1:all}Subtask 1 word-level results by system: The f-measure performance of systems by language; and macro average f-measure of all languages in the last column. Systems marked with * are partial submissions of a specific language set. The performances in bold are best performance of corresponding languages.}
\end{table*}

\section{System Descriptions}
The SIGMORPHON 2022 Shared Task on Morpheme Segmentation received submissions from 7 teams with members from 10 universities and institutes. Many teams submitted more than one system while some focused on a specific set of languages like Romance. In total, we had 24 unique systems over two subtasks, including the baseline system. More system details can be seen in Table~\ref{tab:systems}.

\vspace{1em} \noindent \textbf{AUUH} Researchers at the Aalto University and the University of Helsinki produced six submission systems: two were transformer models and four were bidirectional GRU models created with several innovations of Morfessor feature enrichment, multi-task learning, and multilingual learning. Morfessor \cite{creutz2002unsupervised,creutz2007unsupervised} is the famous language-independent unsupervised and semi-supervised segmentation tool and has a big family of Morfessor variants \cite{virpioja2013morfessor,gronroos2014morfessor,ataman2017linguistically,gronroos2020morfessor}. They have used the first variant of Morfessor \cite{creutz2005unsupervised} for enriching input words along with their Morfessor subword segmentations. AUUH\_A, AUUH\_C, AAUH\_E systems used this Morfessor-based feature enrichment. The key innovation of AUUH systems was multilingual and multi-task traning. They used a similar preprocessing technique \cite{johnson2017google} to distinguish tasks and languages from one another, and then trained multilingual neural models which work on both subtasks. Their transformer-based multilingual and multi-task model, AUUH\_B was the subtask 2 winning system (by its macro average f-measure) and also quite competitive with the subtask 1 winning systems on its partial three-language submissions. 

\vspace{1em} \noindent \textbf{CLUZH} Researchers at the University of Zurich ensembled four submissions \cite{cluzh_sig22} by extending their previous neural hard-attention transducer models \cite{makarov2018uzh,makarov2018imitation,makarov-clematide-2020-cluzh}. For subtask 1, they submit the following strong ensemble \textbf{CLUZH} composed of 3 models without encoder dropout and 2 models with encoder dropout of 0.15. In the sentence-level subtask 2, they submitted three ensembles, and treated this problem as the word-level problem by tokenizing sentences into words. They have also used POS tags as additional features to provide a light for the context of words. All individual models have an encoder dropout probability of 0.25 and vary only in their use of features: \textbf{CLUZH-1} with 3 models without POS features, \textbf{CLUZH-2} with 3 models with POS tag features, and \textbf{CLUZH-3} with combined all the models from CLUZH-1 and CLUZH-2. In overall, the \textbf{CLUZH-3} system was the subtask 2 winning system (by winning two out of three languages) and in subtask 1 \textbf{CLUZH} was the only system, outranked one (DeepSPIN-1) of three DeepSPIN systems. 

\vspace{1em} \noindent \textbf{DeepSPIN} Researchers submitted three neural seq2seq models: (1) \textbf{DeepSPIN-1}, a character-level LSTM with soft attention \cite{bahdanau2014neural} with softmax trained with cross-entropy loss; (2) \textbf{DeepSPIN-2}, a character-level LSTM with soft attention in which softwax is replaced with its sparser version,  1.5-entmax \cite{peters2019sigmorphon}; (3) \textbf{DeepSPIN-3}, a subword-level transformer \cite{vaswani2017attention} with the proposed 1.5- entmax, in which subword segments are  modelled using ULM \cite{kudo-2018-subword}. This design was one of most innovative architectures among all submitted systems. The authors previously experimented with the 1.5-entmax function on other tasks, demonstrating its utility, especially in the tasks with less uncertainty in the search space (e.g., compared to language modelling or machine translation) such as morphological and phonological modelling \cite{peters-martins-2020-one}. The final results of this year's shared task confirm these observations: \textbf{DeepSPIN-2} and \textbf{DeepSPIN-3} achieve superior results and are the winner of the shared task.

\begin{table*}[!h]
\scriptsize
\newcolumntype{R}{>{\raggedleft\arraybackslash}X}
\newcolumntype{C}{>{\centering\arraybackslash}X}
\newcolumntype{L}{>{\raggedright\arraybackslash}X}%%
\centering
\begin{tabularx}{\textwidth}{CCc|lllllll|l}
\hline
inf.                & drv.                & cmp.                & eng        & fra        & ita        & rus        & mon         & hun        & spa        & macro avg. \\
\hline
\hline
\multirow{2}{*}{-}  & \multirow{2}{*}{-}  & \multirow{2}{*}{-}  & \textbf{83.80}      & 84.08      & 82.69*      & 82.56*      & 93.37       & \textbf{85.52}      & 83.58      & 83.6       \\
                     &                      &                      & CLUZH      & DeepSPIN-3 & DeepSPIN-3 & DeepSPIN-1 & JB132       & DeepSPIN-3 & DeepSPIN-2 & DeepSPIN-3 \\
\hline
\hline
\multirow{2}{*}{-}  & \multirow{2}{*}{-}  & \multirow{2}{*}{\checkmark} & 93.23      & \textbf{81.80}      & \textbf{58.10}*      & \textbf{77.67}      & 100.00      & 85.89      & \textbf{57.89}*      & \textbf{78.60}      \\
                     &                      &                      & AUUH\_A    & CLUZH      & CLUZH      & DeepSPIN-2 & all systems & DeepSPIN-3 & DeepSPIN-3 & DeepSPIN-3 \\
\hline
\hline
\multirow{2}{*}{-}  & \multirow{2}{*}{\checkmark} & \multirow{2}{*}{-}  & 94.12      & 87.36*      & 94.62      & 91.4       & \textbf{92.41}       & 94.96      & 92.47      & 92.48      \\
                     &                      &                      & DeepSPIN-3 & DeepSPIN-3 & DeepSPIN-3 & DeepSPIN-3 & DeepSPIN-3  & DeepSPIN-3 & DeepSPIN-3 & DeepSPIN-3 \\
\hline
\hline
\multirow{2}{*}{\checkmark} & \multirow{2}{*}{-}  & \multirow{2}{*}{-}  & 91.29*      & 96.37      & 96.27      & 99.75      & 99.66       & 98.31      & 98.81      & 96.97      \\
                     &                      &                      & CLUZH      & CLUZH      & CLUZH      & DeepSPIN-3 & DeepSPIN-3  & DeepSPIN-3 & DeepSPIN-2 & DeepSPIN-3 \\
\hline
\hline
\multirow{2}{*}{-}  & \multirow{2}{*}{\checkmark} & \multirow{2}{*}{\checkmark} & 95.74      & 80.61      & 70.59*      & 92.13      & -           & 89.82      & 97.3       & 87.65      \\
                     &                      &                      & DeepSPIN-2 & DeepSPIN-3 & DeepSPIN-3 & DeepSPIN-3 & -           & DeepSPIN-3 & DeepSPIN-3 & DeepSPIN-3 \\
\hline
\hline
\multirow{2}{*}{\checkmark} & \multirow{2}{*}{-}  & \multirow{2}{*}{\checkmark} & 96.89      & 96.60      & 94.97      & 100        & 100         & 98.71      & 96.15      & 97.45      \\
                     &                      &                      & DeepSPIN-3 & DeepSPIN-2 & DeepSPIN-3 & DeepSPIN-3 & all systems & DeepSPIN-3 & DeepSPIN-1 & DeepSPIN-3 \\
\hline
\hline
\multirow{2}{*}{\checkmark} & \multirow{2}{*}{\checkmark} & \multirow{2}{*}{-}  & 97.54      & 99.03      & 99.23      & 99.97      & 99.74       & 99.41      & 99.75      & 99.24      \\
                     &                      &                      & DeepSPIN-3 & DeepSPIN-3 & DeepSPIN-3 & DeepSPIN-3 & DeepSPIN-3  & DeepSPIN-2 & DeepSPIN-3 & DeepSPIN-3 \\
\hline
\hline
\multirow{2}{*}{\checkmark} & \multirow{2}{*}{\checkmark} & \multirow{2}{*}{\checkmark} & 97.13      & 100        & 100        & 99.88      & -           & 99.28      & 97.04      & 98.23      \\
                     &                      &                      & DeepSPIN-3 & DeepSPIN-3 & DeepSPIN-2 & DeepSPIN-2 & -           & DeepSPIN-2 & DeepSPIN-2 & DeepSPIN-2 \\
\hline
\end{tabularx}
\caption{\label{tab:besttask1:all}Subtask 1 word-level results by morphological category: f-measure performance of best performing system on a corresponding language and a category; Numbers in bold are worst performance of their corresponding language. Performances marked with * are worst performances of their morphological category.}
\end{table*}

\vspace{1em} \noindent \textbf{GU}  One team from Georgetown University produced two submissions for three Romance languages of the word-level subtask, based on the GRU-based encoder-decoder model \cite{GU2022}. In initial attempts, they tried to use additional features from the Wiktionary lists of prefixes and suffixes to train the model. However, such additional features decreased the main performances across morphological categories, so they excluded these features from the final submissions. Later on, they focus on data sharing between Romance languages. In French, the training data were augmented with four morphological category data from Italian and Spanish training and development datasets. These categories include non-inflection categories of \texttt{000}, \texttt{001}, \texttt{010}, \texttt{011}. With these experiments, they made minor improvements to these three languages. For these results, more research is needed to understand that transfer learning is useful.

\vspace{1em} \noindent \textbf{NUM DI} A single submission from the National University of Mongolia \cite{Task2_NUMDI} is a transformer-based neural model. Their model architecture is simple as single-layered encoder-decoder classic architecture. All the hyper-parameter settings are same as fairseq's standard tutorial tool. Their submission is also limited by four languages of subtask 1 due to human error.

\vspace{1em} \noindent \textbf{JB132} The Charles University team \cite{JB132} designed the Hidden Markov model, trained with the expectation-maximization algorithm. This model architecture has two sub-models. The first sub-model takes words as input and converts them into candidate morphemes. The second sub-model takes candidate morphemes and generates morphs as output. The first sub-model has three generators for accounting prefixes, root words, and suffixes. It is the only system not using neural methods among all submitted systems and the system's prediction is interpretable and can be useful for error analysis.

\vspace{1em} \noindent \textbf{Tü Seg} The University of Tübingen \cite{Task2-TueSeg} team submitted two systems for each of subtasks. Both systems extend the sequence-labeling method proposed by \cite{hellwig2018sanskrit,li2022word}. Their systems are very innovative and unique among all other neural models for considering the main segmentation task as a sequence-labeling task. All other neural systems used seq2seq architecture. Their neural model used a plain two-layer BiLSTM architecture. By its design, Tü Seg systems have at least two advantages over the main seq2seq alternative: (a) the number of parameters is much fewer, so the model can be trained fast and process quickly; (b) the system predictions are more interpretable compared to other neural systems and can help with the error analyses of high-resource datasets.  

\begin{figure*}[!h]
\begin{center}
%\fbox{\parbox{6cm}{
%This is a figure with a caption.}}
\includegraphics[width=\textwidth]{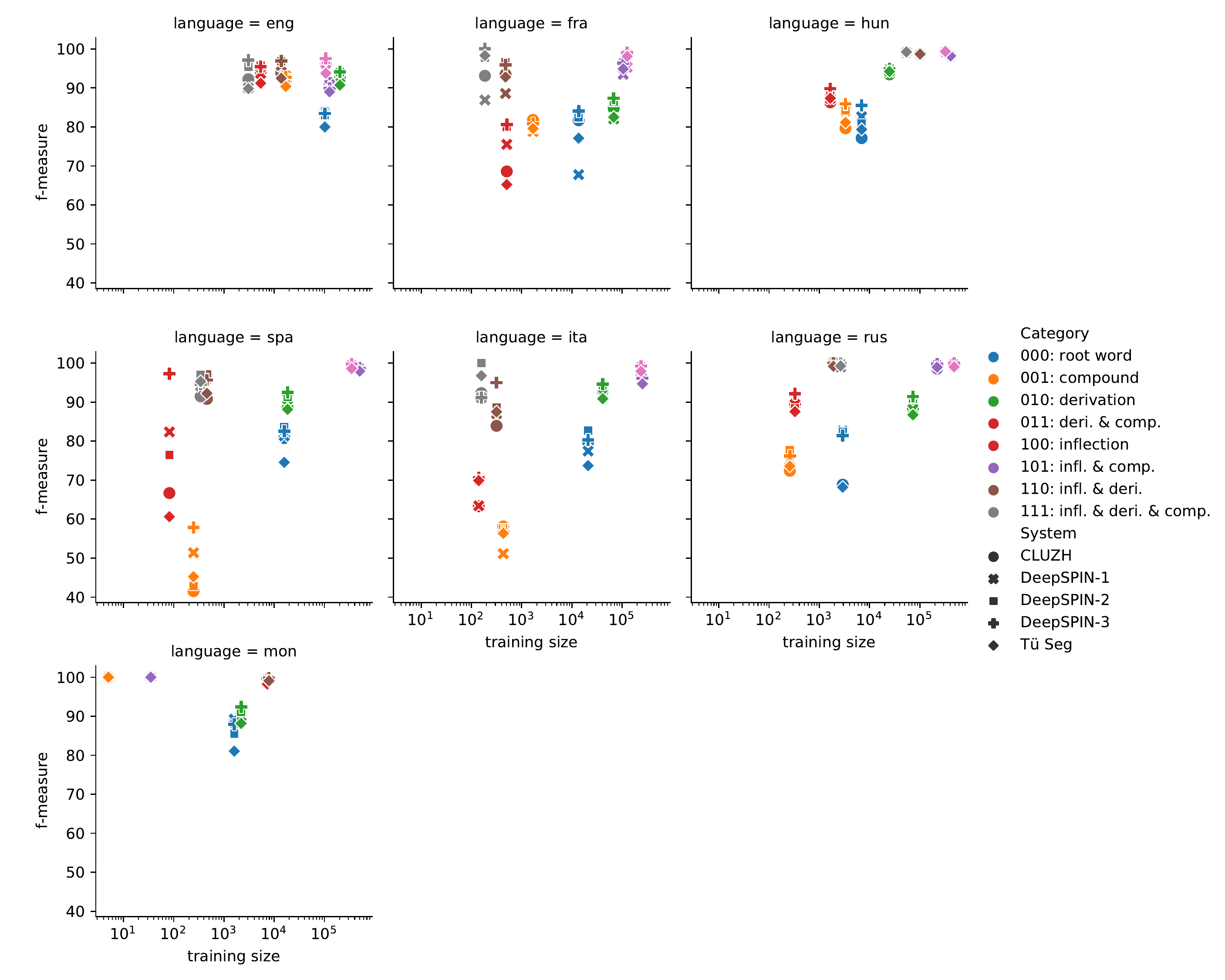}
\caption{Impact of training sizes over languages and morphological categories: Results from top5-ranked systems of word-level subtask 1}
\label{fig:size}
\end{center}
\end{figure*}

\begin{figure*}[!h]
\begin{center}
%\fbox{\parbox{6cm}{
%This is a figure with a caption.}}
\includegraphics[width=\textwidth]{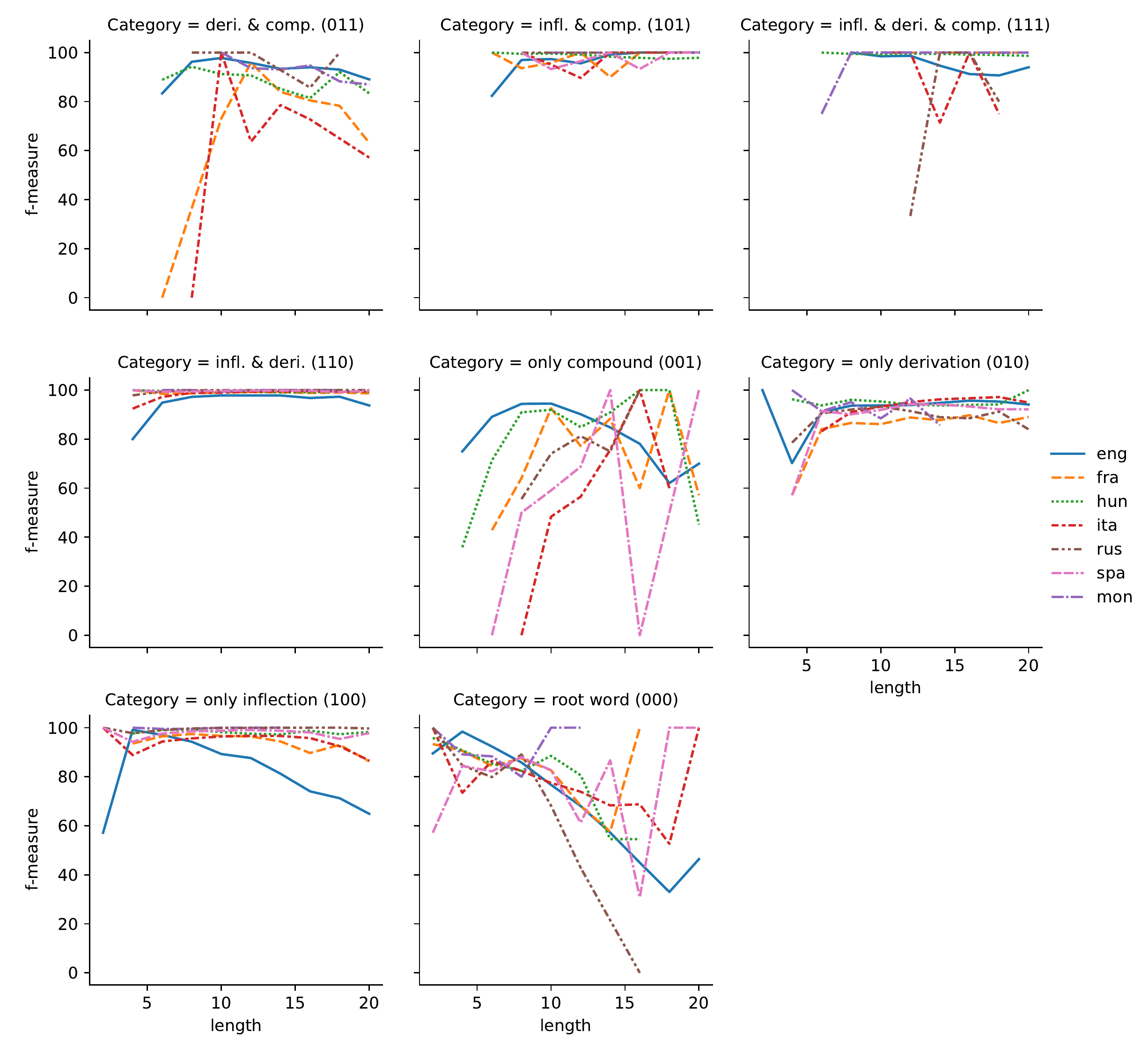}
\caption{Impact of word length over languages and morphological categories: Results from DeepSPIN-3, the winning system of subtask 1, word-level morpheme segmentation}
\label{fig:word_len}
\end{center}
\end{figure*}

\section{The System Results}
All system results can be found and downloaded from the shared task GitHub page.\footnote{\url{https://github.com/sigmorphon/2022SegmentationST/tree/main/results}}

\subsection{Subtask 1 word-level results}
Relative system performance of subtask 1 is provided in Table~\ref{tab:subtask1:all} which shows each system's f-measure by languages. The best performance of each language from submitted systems is in bold. 

Two teams exploited external resources in some form: AUUH and GU. In general, any relative performance gained was minimal. AUUH submitted two systems that used additional resources, they received extra ~1\% compared to the team's other systems. Similarly, GU and their submitted systems saw some minimal improvements over the performances. This details can be seen from their system description paper \cite{GU2022}. 

Only two of all the systems submitted to subtask 1 were multilingual and multi-task learning at same time. These two systems were proposed by AUUH team, but partial-language submissions were for English, Czech, and Mongolian. The important insight from this experiment is that the multi-task and multilingual learning approaches are quite beneficial for the task because their partial performances are quite competitive with the winning systems, DeepSPIN-3, DeepSPIN-2, and CLUZH. 

% Even though the top-ranking systems perform very well on their own, other systems may have some amount of complementary information across morphological categories, so that we listed the best-performed systems of combinations of each language and each morphological category in Table \ref{tab:besttask1:all}. From this table, the performances in bold are lowest performances of the corresponding languages. For instance, Enlgish root words (having 83.80 f-measure) are much hard to predict than other morphological-category words in English. Another observation is that the compound words in general of all languages are least \%5 harder to the second hardest category, root words \texttt{000}. 

\vspace{1em} \noindent \textbf{Impact of training size:} In subtask 1, the training datasets' sizes vary across languages and morphological categories. It might have impacted the top-ranked systems. Therefore, we plotted the top5-ranked systems over training size and f-measure performance across morphological categories, as shown in Figure \ref{fig:size}. Here, in high-resource setting (as greater than $10^5$) in all morphological categories, any of the top5-ranked systems always achieves 80\% f-measure greater than 80\%. 

The root words are present in all types of resources settings from high to low. All the systems in this category of root words achieved no more than 85.5\% f-measure except for Mongolian. 

The two inflectional categories \texttt{100} and \texttt{110} are always in high-resource setting, having more than $10^6$ training instances (except for two low-resource languages Czech and Mongolian). All systems achieved their best system performance over these two categories, compared to other categories.

\vspace{1em} \noindent \textbf{Impact of word length:} 
In many NLP tasks, the length of the input sequence is strongly correlated with the difficulty of their tasks \cite{yin2017comparative,wu2018phrase}. So, we present how the DeepSPIN-3's (subtask 1 winning system) performance relates to the word length across languages and morphological categories. Figure \ref{fig:word_len} shows various related facts: (i) for root words \texttt{000}, overall performance decreases across languages with increasing word length; (ii) inflectional morphology is systematically far more productive than other morphological categories, so this fact is reproduced here: the main inflectional category \texttt{100} has consistently high performance across languages and word lengths. 
\begin{table*}[!h]
\small
\newcolumntype{R}{>{\raggedleft\arraybackslash}X}
\newcolumntype{C}{>{\centering\arraybackslash}X}
\newcolumntype{L}{>{\raggedright\arraybackslash}X}%%
\centering
\begin{tabularx}{\textwidth}{l|RRRr|RRRr|RRRr|rr}
\multirow{2}{*}{System} & \multicolumn{4}{c|}{Czech}     & \multicolumn{4}{c|}{English}   & \multicolumn{4}{c|}{Mongolian} & \multicolumn{2}{c}{Macro avg.} \\
\cline{2-15}
                        & \multicolumn{1}{c}{P}     & \multicolumn{1}{c}{R}     & \multicolumn{1}{c}{$F_1$}    & \multicolumn{1}{c|}{Lev.}  & \multicolumn{1}{c}{P}     & \multicolumn{1}{c}{R}     & \multicolumn{1}{c}{$F_1$}    & \multicolumn{1}{c|}{Lev.}  & \multicolumn{1}{c}{P}     & \multicolumn{1}{c}{R}     & \multicolumn{1}{c}{$F_1$}    & \multicolumn{1}{c|}{Lev.}  & \multicolumn{1}{c}{$F_1$}    & \multicolumn{1}{c}{Lev.}          \\
\hline
\hline
WordPiece                & 38.47 & 31.45 & 34.61 & 17.88 & 62.02 & 65.13 & 63.53 & 5.54  & 19.82 & 29.20 & 23.62 & 29.19 & 40.59          & 17.54         \\
ULM & 41.98 & 30.39 & 35.26 & 16.39 & 62.32 & 69.24 & 65.60 & 5.68 & 38.79 & 35.58 & 37.12 & 20.76 & 45.99 & 14.28 \\
Morfessor2 & 49.89 & 36.95 & 42.45 & 13.09 & 54.61 & 69.75 & 61.25 & 6.00 & 50.88 & 45.91 & 48.26 & 17.16 & 50.65 & 12.08 \\

\hline
\hline
AUUH\_A                 & 89.70 & 87.53 & 88.60 & 4.97  & 96.66 & 95.78 & 96.22 & 1.86  & 83.49 & 80.94 & 82.19 & 5.42  & 89.00          & 4.08          \\
AUUH\_B                 & 91.89 & 89.00 & 90.42 & 3.96  & \textbf{96.82} & \textbf{95.79} & \textbf{96.31} & \textbf{1.39}  & 83.74 & 81.46 & 82.59 & 5.16  & \textbf{89.77}          & \textbf{3.50}          \\
AUUH\_C                 & 50.60 & 69.19 & 58.45 & 71.37 & 84.77 & 71.67 & 77.67 & 19.13 & 79.07 & 73.45 & 76.15 & 17.33 & 70.76          & 35.94         \\
AUUH\_D                 & 45.07 & 67.82 & 54.15 & 80.67 & 93.29 & 83.41 & 88.07 & 10.58 & 77.99 & 74.15 & 76.02 & 17.88 & 72.75          & 36.38         \\
AUUH\_E                 & 57.39 & 67.22 & 61.92 & 55.92 & 95.23 & 76.82 & 85.04 & 12.36 & 73.34 & 72.01 & 72.67 & 24.88 & 73.21          & 31.05         \\
AUUH\_F                 & 62.36 & 43.82 & 51.47 & 61.84 & 91.50 & 74.84 & 82.34 & 13.30 & 75.50 & 59.22 & 66.38 & 33.91 & 66.73          & 36.35         \\
\hline
\hline
CLUZH-1                 & 92.03 & 90.69 & 91.35 & 1.93  & 89.74 & 89.20 & 89.47 & 9.86  & 82.98 & 81.48 & 82.22 & 5.28  & 87.68          & 5.69          \\
CLUZH-2                 & 92.41 & 91.13 & 91.76 & 1.87  & 89.71 & 89.22 & 89.47 & 9.79  & 83.29 & 81.83 & 82.55 & 5.19  & 87.93          & 5.62          \\
CLUZH-3                 & \textbf{92.63} & \textbf{91.35} & \textbf{91.99} & \textbf{1.80}  & 89.83 & 89.25 & 89.54 & 9.84  & \textbf{83.71} & \textbf{82.07} & \textbf{82.88} & \textbf{5.10}  & 88.14          & 5.58          \\
\hline
\hline
Tü\_Seg-2                  & 89.52 & 88.42 & 88.97 & 2.50  & 87.83 & 89.58 & 88.69 & 1.78  & 69.59 & 67.55 & 68.55 & 9.85  & 82.07          & 4.71         
\end{tabularx}
\caption{\label{tab:subtask2:all}Subtask 2 sentence-level results: F-measure across 3 languages}
\end{table*}

\vspace{1em} \noindent \textbf{Difficulty of morphological categories:} Even though the top-ranking systems perform very well on their own, other systems may have some complementary information across morphological categories. Therefore, we listed the best-performing systems for combinations of each language and each morphological category in Table~ \ref{tab:besttask1:all}. In the table, the lowest scores in corresponding languages are provided in bold. For instance, English root words (83.80 f-measure) are much harder to predict than other morphological categories in English. The hardest morphological categories are roots \texttt{000}, compounds \texttt{001}, and derivation and compound words \texttt{011}. The winning system, DeepSPIN-3 (marked with + in Figure~\ref{fig:size}), is consistently winning in these three categories across languages. Another observation from Figure~\ref{fig:word_len} is that compound and root words are getting harder to predict across languages with the increase of word length. Also, identifying inflections from short words (word length~\textless~5) is one of the unsolved challenges in all languages (except for English), as shown in Figure~\ref{fig:word_len}. 

\subsection{Subtask 2 sentence-level results}
Relative system performance is described in Table~\ref{tab:subtask2:all}, showing all four evaluation metrics by each combination of system and language. In the sentence-level subtask 2, we have two winners: CLUZH-3 (won two out of three languages) and AUUH\_B (F1 89.77 as maximum macro- average among submissions).

The performance of systems in the sentence-level subtask significantly decreased by 15\% in Mongolian compared to the results of the word-level subtask. One reason is that all submitted systems treated this problem as a zero-shot solution of word-level subtask 1, and mostly ignored its context by their design. 

% \section{Related Works}
% \citet{cotterell2016joint} provided two formal definitions of morphological segmentation: ``surface'' segmentation and ``canonical'' segmentation. Given word $w$, its ``surface'' segmentation is a sequence of \textit{surface substrings} whose concatenation is $w$, e.g., \textit{funniest} → \textit{funn-i-est}. The purpose of canonical segmentation \cite{Task2-TueSeg,kann-etal-2016-neural} is not only to segment morphologically as surface segmentation but also to restore the standardised forms of morphemes, e.g., \textit{funniest} → \textit{fun-y-est}. %If we read these two definitions from taxonomic perspective, the surface segmentation is a special case of canonical seg mentation. 

% Surface segmentation are represented by two groups of work

% % Recent years are the era of subword tokenizaiton algorithms which became the fundamental parts of all state-of-the-art language models, e.g., 

% \begin{itemize}
% \item Surface Segmentation
% \begin{itemize}
% \item BPE
% \item ULM
% \item Morfessor
% \item LVMR
% \end{itemize}
% \item Canonical Segmentation
% \begin{itemize}
% \item \cite{cotterell2016joint}
% \item \cite{kann-etal-2016-neural}
% \item \cite{kann-etal-2018-fortification}
% \end{itemize}
% \item Segmentation resources
% \begin{itemize}
% \item WikiInflection
% \item Universal Segmentations
% \item Word-Formation Networks
% \item UniMorph, MorphyNet, UniDer, MorphoChallenge, CeLex, any other???
% \end{itemize}
% \end{itemize}

\section{Future Directions}
The submitted systems achieved unexpectedly high accuracy across nine languages. This result suggests that the neural systems may have more capabilities beyond segmenting morphemes. For the next year, we plan to modify the task design and enrich the dataset with more fine-grained analysis. For example, \textit{truckdrivers} → \textit{truck @@drive @@er @@s} → \textit{truck \$\$drive @@er \#\#s} where \$\$ is compound, @@ is derivation, and \#\# is inflection.
In another direction, we will explore possibilities of adapting other morphological resources including word-formation resources \cite{zeller2013derivbase,talamo2016derivatario,derinet-2019,vodolazsky2020derivbase} or segmentation resources, UniSegments \cite{unisegments-lrec-2022,unisegments-data-2022}. Our shared task team welcomes continued contributions from the community.

% \begin{itemize}
% \item more diverse set of languages
% \item low-resource settings of languages are encouraged from participants
% \item increase the task difficulty by design
% \end{itemize}

\section{Conclusion}
The SIGMORPHON 2022 Shared Task on Morpheme Segmentation significantly expanded the problem of morphological segmentation, making it more linguistically plausible. %by accounting canonical segmentation and providing explicit performance evaluation of all morphological categories for segmentations.
In this task, seven teams submitted 23 systems for two subtasks in total of nine languages, achieving at minimum F1 30.71 improvement over the three baselines of the state-of-the-art subword tokenization and morphological segmentation tools, being used to train large language models, e.g., XLNet \cite{yang2019xlnet}. The results suggest many directions for improving morpheme segmentation shared task.

\nocite{Ando2005,borschinger-johnson-2011-particle,andrew2007scalable,rasooli-tetrault-2015,goodman-etal-2016-noise,harper-2014-learning}

\section*{Acknowledgements}
We thank Garrett Nicolai and Eleanor Chodroff for their advice and support. The authors also thank Ben Peters and Simon Clematide for their invaluable contributions and advice, including developing the evaluation tool and early detection of data errors. 

\bibliography{anthology,custom}

\begin{thebibliography}{80}
\expandafter\ifx\csname natexlab\endcsname\relax\def\natexlab#1{#1}\fi

\bibitem[{Ando and Zhang(2005)}]{Ando2005}
Rie~Kubota Ando and Tong Zhang. 2005.
\newblock A framework for learning predictive structures from multiple tasks
  and unlabeled data.
\newblock \emph{Journal of Machine Learning Research}, 6:1817--1853.

\bibitem[{Andrew and Gao(2007)}]{andrew2007scalable}
Galen Andrew and Jianfeng Gao. 2007.
\newblock Scalable training of {L1}-regularized log-linear models.
\newblock In \emph{Proceedings of the 24th International Conference on Machine
  Learning}, pages 33--40.

\bibitem[{Ataman et~al.(2017)Ataman, Negri, Turchi, and
  Federico}]{ataman2017linguistically}
Duygu Ataman, Matteo Negri, Marco Turchi, and Marcello Federico. 2017.
\newblock Linguistically motivated vocabulary reduction for neural machine
  translation from turkish to english.

\bibitem[{Bahdanau et~al.(2014)Bahdanau, Cho, and Bengio}]{bahdanau2014neural}
Dzmitry Bahdanau, Kyunghyun Cho, and Yoshua Bengio. 2014.
\newblock Neural machine translation by jointly learning to align and
  translate.
\newblock \emph{arXiv preprint arXiv:1409.0473}.

\bibitem[{Batsuren et~al.(2019{\natexlab{a}})Batsuren, Bella, and
  Giunchiglia}]{batsuren2019cognet}
Khuyagbaatar Batsuren, G{\'a}bor Bella, and Fausto Giunchiglia.
  2019{\natexlab{a}}.
\newblock Cognet: A large-scale cognate database.
\newblock In \emph{Proceedings of the 57th annual meeting of the association
  for computational linguistics}, pages 3136--3145.

\bibitem[{Batsuren et~al.(2021{\natexlab{a}})Batsuren, Bella, and
  Giunchiglia}]{batsuren2021large}
Khuyagbaatar Batsuren, G{\'a}bor Bella, and Fausto Giunchiglia.
  2021{\natexlab{a}}.
\newblock \href {https://doi.org/10.1007/s10579-021-09544-6} {A large and
  evolving cognate database}.
\newblock \emph{Language Resources and Evaluation}.

\bibitem[{Batsuren et~al.(2021{\natexlab{b}})Batsuren, Bella, and
  Giunchiglia}]{batsuren-etal-2021-morphynet}
Khuyagbaatar Batsuren, G{\'a}bor Bella, and Fausto Giunchiglia.
  2021{\natexlab{b}}.
\newblock \href {https://doi.org/10.18653/v1/2021.sigmorphon-1.5}
  {{M}orphy{N}et: a large multilingual database of derivational and
  inflectional morphology}.
\newblock In \emph{Proceedings of the 18th SIGMORPHON Workshop on Computational
  Research in Phonetics, Phonology, and Morphology}, pages 39--48, Online.
  Association for Computational Linguistics.

\bibitem[{Batsuren et~al.(2019{\natexlab{b}})Batsuren, Ganbold, Chagnaa, and
  Giunchiglia}]{batsuren-etal-2019-building}
Khuyagbaatar Batsuren, Amarsanaa Ganbold, Altangerel Chagnaa, and Fausto
  Giunchiglia. 2019{\natexlab{b}}.
\newblock \href {https://aclanthology.org/2019.gwc-1.30} {Building the
  {M}ongolian {W}ord{N}et}.
\newblock In \emph{Proceedings of the 10th Global Wordnet Conference}, pages
  238--244, Wroclaw, Poland. Global Wordnet Association.

\bibitem[{Batsuren et~al.(2022)Batsuren, Goldman, Khalifa, Habash, Kieraś,
  Bella, Leonard, Nicolai, Gorman, Ate, Ryskina, Mielke, Budianskaya,
  El-Khaissi, Pimentel, Gasser, Lane, Raj, Coler, Samame, Camaiteri, Rojas,
  Francis, Oncevay, Bautista, Villegas, Hennigen, Ek, Guriel, Dirix, Bernardy,
  Scherbakov, Bayyr-ool, Anastasopoulos, Zariquiey, Sheifer, Ganieva, Cruz,
  Karahóǧa, Markantonatou, Pavlidis, Plugaryov, Klyachko, Salehi, Angulo,
  Baxi, Krizhanovsky, Krizhanovskaya, Salesky, Vania, Ivanova, White, Maudslay,
  Valvoda, Zmigrod, Czarnowska, Nikkarinen, Salchak, Bhatt, Straughn, Liu,
  Washington, Pinter, Ataman, Wolinski, Suhardijanto, Yablonskaya, Stoehr,
  Dolatian, Nuriah, Ratan, Tyers, Ponti, Aiton, Arora, Hatcher, Kumar, Young,
  Rodionova, Yemelina, Andrushko, Marchenko, Mashkovtseva, Serova,
  Prud'hommeaux, Nepomniashchaya, Giunchiglia, Chodroff, Hulden, Silfverberg,
  McCarthy, Yarowsky, Cotterell, Tsarfaty, and Vylomova}]{batsuren2022unimorph}
Khuyagbaatar Batsuren, Omer Goldman, Salam Khalifa, Nizar Habash, Witold
  Kieraś, Gábor Bella, Brian Leonard, Garrett Nicolai, Kyle Gorman,
  Yustinus~Ghanggo Ate, Maria Ryskina, Sabrina~J. Mielke, Elena Budianskaya,
  Charbel El-Khaissi, Tiago Pimentel, Michael Gasser, William Lane, Mohit Raj,
  Matt Coler, Jaime Rafael~Montoya Samame, Delio~Siticonatzi Camaiteri,
  Esaú~Zumaeta Rojas, Didier~López Francis, Arturo Oncevay, Juan~López
  Bautista, Gema Celeste~Silva Villegas, Lucas~Torroba Hennigen, Adam Ek, David
  Guriel, Peter Dirix, Jean-Philippe Bernardy, Andrey Scherbakov, Aziyana
  Bayyr-ool, Antonios Anastasopoulos, Roberto Zariquiey, Karina Sheifer, Sofya
  Ganieva, Hilaria Cruz, Ritván Karahóǧa, Stella Markantonatou, George
  Pavlidis, Matvey Plugaryov, Elena Klyachko, Ali Salehi, Candy Angulo, Jatayu
  Baxi, Andrew Krizhanovsky, Natalia Krizhanovskaya, Elizabeth Salesky, Clara
  Vania, Sardana Ivanova, Jennifer White, Rowan~Hall Maudslay, Josef Valvoda,
  Ran Zmigrod, Paula Czarnowska, Irene Nikkarinen, Aelita Salchak, Brijesh
  Bhatt, Christopher Straughn, Zoey Liu, Jonathan~North Washington, Yuval
  Pinter, Duygu Ataman, Marcin Wolinski, Totok Suhardijanto, Anna Yablonskaya,
  Niklas Stoehr, Hossep Dolatian, Zahroh Nuriah, Shyam Ratan, Francis~M. Tyers,
  Edoardo~M. Ponti, Grant Aiton, Aryaman Arora, Richard~J. Hatcher, Ritesh
  Kumar, Jeremiah Young, Daria Rodionova, Anastasia Yemelina, Taras Andrushko,
  Igor Marchenko, Polina Mashkovtseva, Alexandra Serova, Emily Prud'hommeaux,
  Maria Nepomniashchaya, Fausto Giunchiglia, Eleanor Chodroff, Mans Hulden,
  Miikka Silfverberg, Arya~D. McCarthy, David Yarowsky, Ryan Cotterell, Reut
  Tsarfaty, and Ekaterina Vylomova. 2022.
\newblock \href {https://doi.org/10.48550/ARXIV.2205.03608} {Unimorph 4.0:
  Universal morphology}.

\bibitem[{Bodnár(2022)}]{JB132}
Jan Bodnár. 2022.
\newblock Jb132 submission to the sigmorphon 2022 shared task 3 on
  morphological segmentation.
\newblock In \emph{19th SIGMORPHON Workshop on Computational Research in
  Phonetics, Phonology, and Morphology}.

\bibitem[{Brown et~al.(2020)Brown, Mann, Ryder, Subbiah, Kaplan, Dhariwal,
  Neelakantan, Shyam, Sastry, Askell et~al.}]{brown2020language}
Tom Brown, Benjamin Mann, Nick Ryder, Melanie Subbiah, Jared~D Kaplan, Prafulla
  Dhariwal, Arvind Neelakantan, Pranav Shyam, Girish Sastry, Amanda Askell,
  et~al. 2020.
\newblock Language models are few-shot learners.
\newblock \emph{Advances in neural information processing systems},
  33:1877--1901.

\bibitem[{Cotterell et~al.(2016)Cotterell, Vieira, and
  Sch{\"u}tze}]{cotterell2016joint}
Ryan Cotterell, Tim Vieira, and Hinrich Sch{\"u}tze. 2016.
\newblock A joint model of orthography and morphological segmentation.
\newblock Association for Computational Linguistics.

\bibitem[{Creutz and Lagus(2002)}]{creutz2002unsupervised}
Mathias Creutz and Krista Lagus. 2002.
\newblock Unsupervised discovery of morphemes.
\newblock In \emph{Proceedings of the ACL-02 Workshop on Morphological and
  Phonological Learning}, pages 21--30.

\bibitem[{Creutz and Lagus(2005)}]{creutz2005unsupervised}
Mathias Creutz and Krista Lagus. 2005.
\newblock \emph{Unsupervised morpheme segmentation and morphology induction
  from text corpora using Morfessor 1.0}.
\newblock Helsinki University of Technology Helsinki.

\bibitem[{Creutz and Lagus(2007)}]{creutz2007unsupervised}
Mathias Creutz and Krista Lagus. 2007.
\newblock Unsupervised models for morpheme segmentation and morphology
  learning.
\newblock \emph{ACM Transactions on Speech and Language Processing (TSLP)},
  4(1):1--34.

\bibitem[{de~Marneffe et~al.(2021)de~Marneffe, Manning, Nivre, and
  Zeman}]{ud-cl-2021}
Marie-Catherine de~Marneffe, Christopher Manning, Joakim Nivre, and Daniel
  Zeman. 2021.
\newblock Universal dependencies.
\newblock \emph{Computational Linguistics}, 47(2):255--308.

\bibitem[{Devlin et~al.(2019)Devlin, Chang, Lee, and
  Toutanova}]{devlin-etal-2019-bert}
Jacob Devlin, Ming-Wei Chang, Kenton Lee, and Kristina Toutanova. 2019.
\newblock \href {https://doi.org/10.18653/v1/N19-1423} {{BERT}: Pre-training of
  deep bidirectional transformers for language understanding}.
\newblock In \emph{Proceedings of the 2019 Conference of the North {A}merican
  Chapter of the Association for Computational Linguistics: Human Language
  Technologies, Volume 1 (Long and Short Papers)}, pages 4171--4186,
  Minneapolis, Minnesota. Association for Computational Linguistics.

\bibitem[{Domingo et~al.(2018)Domingo, Garc{\i}a-Mart{\i}nez, Helle,
  Casacuberta, and Herranz}]{domingo2018much}
Miguel Domingo, Mercedes Garc{\i}a-Mart{\i}nez, Alexandre Helle, Francisco
  Casacuberta, and Manuel Herranz. 2018.
\newblock How much does tokenization affect neural machine translation?
\newblock \emph{arXiv preprint arXiv:1812.08621}.

\bibitem[{Fourrier and Sagot(2020)}]{fourrier2020methodological}
Cl{\'e}mentine Fourrier and Beno{\^\i}t Sagot. 2020.
\newblock Methodological aspects of developing and managing an etymological
  lexical resource: Introducing etymdb 2.0.
\newblock In \emph{LREC 2020-12th Language Resources and Evaluation
  Conference}.

\bibitem[{Gage(1994)}]{gage1994new}
Philip Gage. 1994.
\newblock A new algorithm for data compression.
\newblock \emph{C Users Journal}, 12(2):23--38.

\bibitem[{Girrbach(2022)}]{Task2-TueSeg}
Leander Girrbach. 2022.
\newblock Sigmorphon 2022 shared task on morpheme segmentation submission
  description: Sequence labelling for word-level morpheme segmentation.
\newblock In \emph{19th SIGMORPHON Workshop on Computational Research in
  Phonetics, Phonology, and Morphology}.

\bibitem[{Gow-Smith et~al.(2022)Gow-Smith, Madabushi, Scarton, and
  Villavicencio}]{gow2022improving}
Edward Gow-Smith, Harish~Tayyar Madabushi, Carolina Scarton, and Aline
  Villavicencio. 2022.
\newblock Improving tokenisation by alternative treatment of spaces.
\newblock \emph{arXiv preprint arXiv:2204.04058}.

\bibitem[{Gr{\"o}nroos et~al.(2020)Gr{\"o}nroos, Virpioja, and
  Kurimo}]{gronroos2020morfessor}
Stig-Arne Gr{\"o}nroos, Sami Virpioja, and Mikko Kurimo. 2020.
\newblock Morfessor em+ prune: Improved subword segmentation with expectation
  maximization and pruning.
\newblock \emph{arXiv preprint arXiv:2003.03131}.

\bibitem[{Gr{\"o}nroos et~al.(2014)Gr{\"o}nroos, Virpioja, Smit, and
  Kurimo}]{gronroos2014morfessor}
Stig-Arne Gr{\"o}nroos, Sami Virpioja, Peter Smit, and Mikko Kurimo. 2014.
\newblock Morfessor flatcat: An hmm-based method for unsupervised and
  semi-supervised learning of morphology.
\newblock In \emph{Proceedings of COLING 2014, the 25th International
  Conference on Computational Linguistics: Technical Papers}, pages 1177--1185.

\bibitem[{He et~al.(2020)He, Haffari, and Norouzi}]{he2020dynamic}
Xuanli He, Gholamreza Haffari, and Mohammad Norouzi. 2020.
\newblock Dynamic programming encoding for subword segmentation in neural
  machine translation.
\newblock In \emph{Proceedings of the 58th Annual Meeting of the Association
  for Computational Linguistics}, pages 3042--3051.

\bibitem[{Hellwig and Nehrdich(2018)}]{hellwig2018sanskrit}
Oliver Hellwig and Sebastian Nehrdich. 2018.
\newblock Sanskrit word segmentation using character-level recurrent and
  convolutional neural networks.
\newblock In \emph{Proceedings of the 2018 conference on empirical methods in
  natural language processing}, pages 2754--2763.

\bibitem[{Hiraoka et~al.(2021)Hiraoka, Takase, Uchiumi, Keyaki, and
  Okazaki}]{hiraoka2021joint}
Tatsuya Hiraoka, Sho Takase, Kei Uchiumi, Atsushi Keyaki, and Naoaki Okazaki.
  2021.
\newblock Joint optimization of tokenization and downstream model.
\newblock In \emph{Findings of the Association for Computational Linguistics:
  ACL-IJCNLP 2021}, pages 244--255.

\bibitem[{Johnson et~al.(2017)Johnson, Schuster, Le, Krikun, Wu, Chen, Thorat,
  Vi{\'e}gas, Wattenberg, Corrado et~al.}]{johnson2017google}
Melvin Johnson, Mike Schuster, Quoc~V Le, Maxim Krikun, Yonghui Wu, Zhifeng
  Chen, Nikhil Thorat, Fernanda Vi{\'e}gas, Martin Wattenberg, Greg Corrado,
  et~al. 2017.
\newblock Google’s multilingual neural machine translation system: Enabling
  zero-shot translation.
\newblock \emph{Transactions of the Association for Computational Linguistics},
  5:339--351.

\bibitem[{Kann et~al.(2016)Kann, Cotterell, and
  Sch{\"u}tze}]{kann-etal-2016-neural}
Katharina Kann, Ryan Cotterell, and Hinrich Sch{\"u}tze. 2016.
\newblock \href {https://doi.org/10.18653/v1/D16-1097} {Neural morphological
  analysis: Encoding-decoding canonical segments}.
\newblock In \emph{Proceedings of the 2016 Conference on Empirical Methods in
  Natural Language Processing}, pages 961--967, Austin, Texas. Association for
  Computational Linguistics.

\bibitem[{Kann et~al.(2018)Kann, Mager~Hois, Meza-Ruiz, and
  Sch{\"u}tze}]{kann-etal-2018-fortification}
Katharina Kann, Jesus~Manuel Mager~Hois, Ivan~Vladimir Meza-Ruiz, and Hinrich
  Sch{\"u}tze. 2018.
\newblock \href {https://doi.org/10.18653/v1/N18-1005} {Fortification of neural
  morphological segmentation models for polysynthetic minimal-resource
  languages}.
\newblock In \emph{Proceedings of the 2018 Conference of the North {A}merican
  Chapter of the Association for Computational Linguistics: Human Language
  Technologies, Volume 1 (Long Papers)}, pages 47--57, New Orleans, Louisiana.
  Association for Computational Linguistics.

\bibitem[{Kudo(2018)}]{kudo-2018-subword}
Taku Kudo. 2018.
\newblock \href {https://doi.org/10.18653/v1/P18-1007} {Subword regularization:
  Improving neural network translation models with multiple subword
  candidates}.
\newblock In \emph{Proceedings of the 56th Annual Meeting of the Association
  for Computational Linguistics (Volume 1: Long Papers)}, pages 66--75,
  Melbourne, Australia. Association for Computational Linguistics.

\bibitem[{Kurimo et~al.(2007)Kurimo, Creutz, and
  Varjokallio}]{kurimo2007unsupervised}
Mikko Kurimo, Mathias Creutz, and Matti Varjokallio. 2007.
\newblock Unsupervised morpheme analysis evaluation by a comparison to a
  linguistic gold standard-morpho challenge 2007.
\newblock In \emph{CLEF (Working Notes)}.

\bibitem[{Kurimo et~al.(2008)Kurimo, Turunen, and
  Varjokallio}]{kurimo2008overview}
Mikko Kurimo, Ville Turunen, and Matti Varjokallio. 2008.
\newblock Overview of morpho challenge 2008.
\newblock In \emph{Workshop of the Cross-Language Evaluation Forum for European
  Languages}, pages 951--966. Springer.

\bibitem[{Kurimo et~al.(2010{\natexlab{a}})Kurimo, Virpioja, Turunen, and
  Lagus}]{kurimo2010morpho}
Mikko Kurimo, Sami Virpioja, Ville Turunen, and Krista Lagus.
  2010{\natexlab{a}}.
\newblock Morpho challenge 2005-2010: Evaluations and results.
\newblock In \emph{Proceedings of the 11th Meeting of the ACL Special Interest
  Group on Computational Morphology and Phonology}, pages 87--95.

\bibitem[{Kurimo et~al.(2009)Kurimo, Virpioja, Turunen, Blackwood, and
  Byrne}]{kurimo2009overview}
Mikko Kurimo, Sami Virpioja, Ville~T Turunen, Graeme~W Blackwood, and William
  Byrne. 2009.
\newblock Overview and results of morpho challenge 2009.
\newblock In \emph{Workshop of the Cross-Language Evaluation Forum for European
  Languages}, pages 578--597. Springer.

\bibitem[{Kurimo et~al.(2010{\natexlab{b}})Kurimo, Virpioja, Turunen
  et~al.}]{kurimo2010proceedings}
Mikko Kurimo, Sami Virpioja, Ville~T Turunen, et~al. 2010{\natexlab{b}}.
\newblock Proceedings of the morpho challenge 2010 workshop.
\newblock In \emph{Morpho Challenge Workshop; 2010; Espoo}. Aalto University
  School of Science and Technology.

\bibitem[{Levine(2022)}]{GU2022}
Lauren Levine. 2022.
\newblock Sharing data by language family: Data augmentation for romance
  language morpheme segmentation.
\newblock In \emph{19th SIGMORPHON Workshop on Computational Research in
  Phonetics, Phonology, and Morphology}.

\bibitem[{Li and Girrbach(2022)}]{li2022word}
Jingwen Li and Leander Girrbach. 2022.
\newblock Word segmentation and morphological parsing for sanskrit.
\newblock \emph{arXiv preprint arXiv:2201.12833}.

\bibitem[{Liu et~al.(2019)Liu, Ott, Goyal, Du, Joshi, Chen, Levy, Lewis,
  Zettlemoyer, and Stoyanov}]{liu2019roberta}
Yinhan Liu, Myle Ott, Naman Goyal, Jingfei Du, Mandar Joshi, Danqi Chen, Omer
  Levy, Mike Lewis, Luke Zettlemoyer, and Veselin Stoyanov. 2019.
\newblock Roberta: A robustly optimized bert pretraining approach.
\newblock \emph{arXiv preprint arXiv:1907.11692}.

\bibitem[{Mach{\'a}{\v{c}}ek et~al.(2018)Mach{\'a}{\v{c}}ek, Vidra, and
  Bojar}]{machavcek2018morphological}
Dominik Mach{\'a}{\v{c}}ek, Jon{\'a}{\v{s}} Vidra, and Ond{\v{r}}ej Bojar.
  2018.
\newblock Morphological and language-agnostic word segmentation for nmt.
\newblock In \emph{International Conference on Text, Speech, and Dialogue},
  pages 277--284. Springer.

\bibitem[{Makarov and Clematide(2018{\natexlab{a}})}]{makarov2018imitation}
Peter Makarov and Simon Clematide. 2018{\natexlab{a}}.
\newblock Imitation learning for neural morphological string transduction.
\newblock \emph{arXiv preprint arXiv:1808.10701}.

\bibitem[{Makarov and Clematide(2018{\natexlab{b}})}]{makarov2018uzh}
Peter Makarov and Simon Clematide. 2018{\natexlab{b}}.
\newblock Uzh at conll-sigmorphon 2018 shared task on universal morphological
  reinflection.
\newblock Association for Computational Linguistics.

\bibitem[{Makarov and Clematide(2020)}]{makarov-clematide-2020-cluzh}
Peter Makarov and Simon Clematide. 2020.
\newblock \href {https://doi.org/10.18653/v1/2020.sigmorphon-1.19} {{CLUZH} at
  {SIGMORPHON} 2020 shared task on multilingual grapheme-to-phoneme
  conversion}.
\newblock In \emph{Proceedings of the 17th SIGMORPHON Workshop on Computational
  Research in Phonetics, Phonology, and Morphology}, pages 171--176, Online.
  Association for Computational Linguistics.

\bibitem[{Matthews et~al.(2018)Matthews, Neubig, and Dyer}]{matthews2018using}
Austin Matthews, Graham Neubig, and Chris Dyer. 2018.
\newblock Using morphological knowledge in open-vocabulary neural language
  models.
\newblock In \emph{Proceedings of the 2018 Conference of the North American
  Chapter of the Association for Computational Linguistics: Human Language
  Technologies, Volume 1 (Long Papers)}, pages 1435--1445.

\bibitem[{Mielke et~al.(2021)Mielke, Alyafeai, Salesky, Raffel, Dey, Gall{\'e},
  Raja, Si, Lee, Sagot et~al.}]{mielke2021between}
Sabrina~J Mielke, Zaid Alyafeai, Elizabeth Salesky, Colin Raffel, Manan Dey,
  Matthias Gall{\'e}, Arun Raja, Chenglei Si, Wilson~Y Lee, Beno{\^\i}t Sagot,
  et~al. 2021.
\newblock Between words and characters: A brief history of open-vocabulary
  modeling and tokenization in nlp.
\newblock \emph{arXiv preprint arXiv:2112.10508}.

\bibitem[{Munkhjargal et~al.(2016)Munkhjargal, Chagnaa, and
  Jaimai}]{munkhjargal2016morphological}
Zoljargal Munkhjargal, Altangerel Chagnaa, and Purev Jaimai. 2016.
\newblock Morphological transducer for mongolian.
\newblock In \emph{International Conference on Computational Collective
  Intelligence}, pages 546--554. Springer.

\bibitem[{Nzeyimana and Rubungo(2022)}]{nzeyimana2022kinyabert}
Antoine Nzeyimana and Andre~Niyongabo Rubungo. 2022.
\newblock Kinyabert: a morphology-aware kinyarwanda language model.
\newblock \emph{arXiv preprint arXiv:2203.08459}.

\bibitem[{Pelegrinov{\'a} et~al.(2021)Pelegrinov{\'a}, El{\v s}{\'{\i}}k, {\v
  C}ech, and Ma{\v c}utek}]{morfoczech-data-2022}
Kate{\v r}ina Pelegrinov{\'a}, Viktor El{\v s}{\'{\i}}k, Radek {\v C}ech, and
  J{\'a}n Ma{\v c}utek. 2021.
\newblock \href {http://hdl.handle.net/11234/1-4626} {{MorfoCzech}}.
\newblock {LINDAT}/{CLARIAH}-{CZ} digital library at the Institute of Formal
  and Applied Linguistics ({{\'U}FAL}), Faculty of Mathematics and Physics,
  Charles University.

\bibitem[{Peters and Martins(2020)}]{peters-martins-2020-one}
Ben Peters and Andr{\'e} F.~T. Martins. 2020.
\newblock \href {https://doi.org/10.18653/v1/2020.sigmorphon-1.4}
  {One-size-fits-all multilingual models}.
\newblock In \emph{Proceedings of the 17th SIGMORPHON Workshop on Computational
  Research in Phonetics, Phonology, and Morphology}, pages 63--69, Online.
  Association for Computational Linguistics.

\bibitem[{Peters and Martins(2022)}]{DeepSPIN2022}
Ben Peters and Andr{\'e} F.~T. Martins. 2022.
\newblock Beyond characters: Subword-level morpheme segmentation.
\newblock In \emph{19th SIGMORPHON Workshop on Computational Research in
  Phonetics, Phonology, and Morphology}.

\bibitem[{Peters and Martins(2019)}]{peters2019sigmorphon}
Ben Peters and Andr{\'e}~FT Martins. 2019.
\newblock {IT--IST} at the sigmorphon 2019 shared task: Sparse two-headed
  models for inflection.
\newblock In \emph{Proceedings of the 16th Workshop on Computational Research
  in Phonetics, Phonology, and Morphology}, pages 50--56.

\bibitem[{Provilkov et~al.(2019)Provilkov, Emelianenko, and
  Voita}]{provilkov2019bpe}
Ivan Provilkov, Dmitrii Emelianenko, and Elena Voita. 2019.
\newblock Bpe-dropout: Simple and effective subword regularization.
\newblock \emph{arXiv preprint arXiv:1910.13267}.

\bibitem[{Radford et~al.(2019)Radford, Wu, Child, Luan, Amodei, Sutskever
  et~al.}]{radford2019language}
Alec Radford, Jeffrey Wu, Rewon Child, David Luan, Dario Amodei, Ilya
  Sutskever, et~al. 2019.
\newblock Language models are unsupervised multitask learners.
\newblock \emph{OpenAI blog}, 1(8):9.

\bibitem[{Rasooli and Tetreault(2015)}]{rasooli-tetrault-2015}
Mohammad~Sadegh Rasooli and Joel~R. Tetreault. 2015.
\newblock \href {http://arxiv.org/abs/1503.06733} {Yara parser: {A} fast and
  accurate dependency parser}.
\newblock \emph{Computing Research Repository}, arXiv:1503.06733.
\newblock Version 2.

\bibitem[{Rouhe et~al.(2022)Rouhe, Grönroos, Virpioja, Creutz, and
  Kurimo}]{auuh22sigmorphon}
Aku Rouhe, Stig-Arne Grönroos, Sami Virpioja, Mathias Creutz, and Mikko
  Kurimo. 2022.
\newblock Morfessor-enriched features and multilingual training for canonical
  morphological segmentation.
\newblock In \emph{19th SIGMORPHON Workshop on Computational Research in
  Phonetics, Phonology, and Morphology}.

\bibitem[{Saleva and Lignos(2021)}]{saleva-lignos-2021-effectiveness}
Jonne Saleva and Constantine Lignos. 2021.
\newblock \href {https://doi.org/10.18653/v1/2021.eacl-srw.22} {The
  effectiveness of morphology-aware segmentation in low-resource neural machine
  translation}.
\newblock In \emph{Proceedings of the 16th Conference of the European Chapter
  of the Association for Computational Linguistics: Student Research Workshop},
  pages 164--174, Online. Association for Computational Linguistics.

\bibitem[{Schuster and Nakajima(2012)}]{schuster2012japanese}
Mike Schuster and Kaisuke Nakajima. 2012.
\newblock Japanese and korean voice search.
\newblock In \emph{2012 IEEE international conference on acoustics, speech and
  signal processing (ICASSP)}, pages 5149--5152. IEEE.

\bibitem[{Schwartz et~al.(2020)Schwartz, Tyers, Levin, Kirov, Littell, Lo,
  Prud'hommeaux, Park, Steimel, Knowles et~al.}]{schwartz2020neural}
Lane Schwartz, Francis Tyers, Lori Levin, Christo Kirov, Patrick Littell,
  Chi-kiu Lo, Emily Prud'hommeaux, Hyunji~Hayley Park, Kenneth Steimel, Rebecca
  Knowles, et~al. 2020.
\newblock Neural polysynthetic language modelling.
\newblock \emph{arXiv preprint arXiv:2005.05477}.

\bibitem[{Sennrich et~al.(2016)Sennrich, Haddow, and
  Birch}]{sennrich2016neural}
Rico Sennrich, Barry Haddow, and Alexandra Birch. 2016.
\newblock Neural machine translation of rare words with subword units.
\newblock In \emph{Proceedings of the 54th Annual Meeting of the Association
  for Computational Linguistics (Volume 1: Long Papers)}, pages 1715--1725.

\bibitem[{S{\'e}rasset(2015)}]{serasset2015dbnary}
Gilles S{\'e}rasset. 2015.
\newblock Dbnary: Wiktionary as a lemon-based multilingual lexical resource in
  rdf.
\newblock \emph{Semantic Web}, 6(4):355--361.

\bibitem[{Silveira et~al.(2014)Silveira, Dozat, de~Marneffe, Bowman, Connor,
  Bauer, and Manning}]{silveira14gold}
Natalia Silveira, Timothy Dozat, Marie-Catherine de~Marneffe, Samuel Bowman,
  Miriam Connor, John Bauer, and Christopher~D. Manning. 2014.
\newblock A gold standard dependency corpus for {E}nglish.
\newblock In \emph{Proceedings of the Ninth International Conference on
  Language Resources and Evaluation (LREC-2014)}.

\bibitem[{Slav{\'{\i}}{\v c}kov{\'a} et~al.(2017)Slav{\'{\i}}{\v c}kov{\'a},
  Hlav{\'a}{\v c}ov{\'a}, and Pognan}]{slavickova-2017}
Eleonora Slav{\'{\i}}{\v c}kov{\'a}, Jaroslava Hlav{\'a}{\v c}ov{\'a}, and
  Patrice Pognan. 2017.
\newblock \href {http://hdl.handle.net/11234/1-2546} {Retrograde morphemic
  dictionary of czech - verbs}.
\newblock {LINDAT}/{CLARIAH}-{CZ} digital library at the Institute of Formal
  and Applied Linguistics ({{\'U}FAL}), Faculty of Mathematics and Physics,
  Charles University.

\bibitem[{Song et~al.(2022)Song, Dabre, Chu, Kurohashi, and
  Sumita}]{song2022self}
Haiyue Song, Raj Dabre, Chenhui Chu, Sadao Kurohashi, and Eiichiro Sumita.
  2022.
\newblock Self-supervised dynamic programming encoding for neural machine
  translation.

\bibitem[{Talamo et~al.(2016)Talamo, Celata, and
  Bertinetto}]{talamo2016derivatario}
Luigi Talamo, Chiara Celata, and Pier~Marco Bertinetto. 2016.
\newblock Derivatario: An annotated lexicon of italian derivatives.
\newblock \emph{Word Structure}, 9(1):72--102.

\bibitem[{Tay et~al.(2021)Tay, Tran, Ruder, Gupta, Chung, Bahri, Qin,
  Baumgartner, Yu, and Metzler}]{tay2021charformer}
Yi~Tay, Vinh~Q Tran, Sebastian Ruder, Jai Gupta, Hyung~Won Chung, Dara Bahri,
  Zhen Qin, Simon Baumgartner, Cong Yu, and Donald Metzler. 2021.
\newblock Charformer: Fast character transformers via gradient-based subword
  tokenization.
\newblock \emph{arXiv preprint arXiv:2106.12672}.

\bibitem[{Vaswani et~al.(2017)Vaswani, Shazeer, Parmar, Uszkoreit, Jones,
  Gomez, Kaiser, and Polosukhin}]{vaswani2017attention}
Ashish Vaswani, Noam Shazeer, Niki Parmar, Jakob Uszkoreit, Llion Jones,
  Aidan~N Gomez, {\L}ukasz Kaiser, and Illia Polosukhin. 2017.
\newblock Attention is all you need.
\newblock \emph{Advances in neural information processing systems}, 30.

\bibitem[{Vidra et~al.(2019)Vidra, Žabokrtský, Ševčíková, and
  Kyjánek}]{derinet-2019}
Jonáš Vidra, Zdeněk Žabokrtský, Magda Ševčíková, and Lukáš Kyjánek.
  2019.
\newblock Derinet 2.0: Towards an all-in-one word-formation resource.
\newblock In \emph{Proceedings of the Second International Workshop on
  Resources and Tools for Derivational Morphology (DeriMo 2019)}, pages 81--89,
  Praha, Czechia. ÚFAL MFF UK.

\bibitem[{Virpioja et~al.(2013)Virpioja, Smit, Gr{\"o}nroos, Kurimo
  et~al.}]{virpioja2013morfessor}
Sami Virpioja, Peter Smit, Stig-Arne Gr{\"o}nroos, Mikko Kurimo, et~al. 2013.
\newblock Morfessor 2.0: Python implementation and extensions for morfessor
  baseline.

\bibitem[{Vodolazsky(2020)}]{vodolazsky2020derivbase}
Daniil Vodolazsky. 2020.
\newblock Derivbase. ru: A derivational morphology resource for russian.
\newblock In \emph{Proceedings of The 12th Language Resources and Evaluation
  Conference}, pages 3937--3943.

\bibitem[{Wang et~al.(2021)Wang, Ruder, and Neubig}]{wang2021multi}
Xinyi Wang, Sebastian Ruder, and Graham Neubig. 2021.
\newblock Multi-view subword regularization.
\newblock In \emph{Proceedings of the 2021 Conference of the North American
  Chapter of the Association for Computational Linguistics: Human Language
  Technologies}, pages 473--482.

\bibitem[{Wehrli et~al.(2022)Wehrli, Clematide, and Makarov}]{cluzh_sig22}
Silvan Wehrli, Simon Clematide, and Peter Makarov. 2022.
\newblock Cluzh at sigmorphon 2022 shared tasks on morpheme segmentation and
  inflection generation.
\newblock In \emph{19th SIGMORPHON Workshop on Computational Research in
  Phonetics, Phonology, and Morphology}.

\bibitem[{Wu et~al.(2018)Wu, Wang, Liu, and Ma}]{wu2018phrase}
Wei Wu, Houfeng Wang, Tianyu Liu, and Shuming Ma. 2018.
\newblock Phrase-level self-attention networks for universal sentence encoding.
\newblock In \emph{Proceedings of the 2018 Conference on Empirical Methods in
  Natural Language Processing}, pages 3729--3738.

\bibitem[{Wu and Yarowsky(2020)}]{wu-yarowsky-2020-computational}
Winston Wu and David Yarowsky. 2020.
\newblock \href {https://www.aclweb.org/anthology/2020.lrec-1.397}
  {Computational etymology and word emergence}.
\newblock In \emph{Proceedings of the 12th Language Resources and Evaluation
  Conference}, pages 3252--3259, Marseille, France. European Language Resources
  Association.

\bibitem[{Yang et~al.(2019)Yang, Dai, Yang, Carbonell, Salakhutdinov, and
  Le}]{yang2019xlnet}
Zhilin Yang, Zihang Dai, Yiming Yang, Jaime Carbonell, Russ~R Salakhutdinov,
  and Quoc~V Le. 2019.
\newblock Xlnet: Generalized autoregressive pretraining for language
  understanding.
\newblock \emph{Advances in neural information processing systems}, 32.

\bibitem[{Yin et~al.(2017)Yin, Kann, Yu, and Sch{\"u}tze}]{yin2017comparative}
Wenpeng Yin, Katharina Kann, Mo~Yu, and Hinrich Sch{\"u}tze. 2017.
\newblock Comparative study of cnn and rnn for natural language processing.
\newblock \emph{arXiv preprint arXiv:1702.01923}.

\bibitem[{{\v Z}abokrtsk{\'y} et~al.(2022){\v Z}abokrtsk{\'y}, Bafna,
  Bodn{\'a}r, Kyj{\'a}nek, Svoboda, {\v S}ev{\v c}{\'{\i}}kov{\'a}, Vidra,
  Angle, Ansari, Arkhangelskiy, Batsuren, Bella, Bertinetto, Bonami, Celata,
  Daniel, Fedorenko, Filko, Giunchiglia, Haghdoost, Hathout, Khomchenkova,
  Khurshudyan, Levonian, Litta, Medvedeva, Muralikrishna, Namer, Nikravesh,
  Pad{\'o}, Passarotti, Plungian, Polyakov, Potapov, Pruthwik, Rao~B, Rubakov,
  Samar, Sharma, {\v S}najder, {\v S}ojat, {\v S}tefanec, Talamo, Tribout,
  Vodolazsky, Vydrin, Zakirova, and Zeller}]{unisegments-data-2022}
Zden{\v e}k {\v Z}abokrtsk{\'y}, Nyati Bafna, Jan Bodn{\'a}r, Luk{\'a}{\v s}
  Kyj{\'a}nek, Emil Svoboda, Magda {\v S}ev{\v c}{\'{\i}}kov{\'a}, Jon{\'a}{\v
  s} Vidra, Sachi Angle, Ebrahim Ansari, Timofey Arkhangelskiy, Khuyagbaatar
  Batsuren, G{\'a}bor Bella, Pier~Marco Bertinetto, Olivier Bonami, Chiara
  Celata, Michael Daniel, Alexei Fedorenko, Matea Filko, Fausto Giunchiglia,
  Hamid Haghdoost, Nabil Hathout, Irina Khomchenkova, Victoria Khurshudyan,
  Dmitri Levonian, Eleonora Litta, Maria Medvedeva, S.~N. Muralikrishna,
  Fiammetta Namer, Mahshid Nikravesh, Sebastian Pad{\'o}, Marco Passarotti,
  Vladimir Plungian, Alexey Polyakov, Mihail Potapov, Mishra Pruthwik, Ashwath
  Rao~B, Sergei Rubakov, Husain Samar, Dipti~Misra Sharma, Jan {\v S}najder,
  Kre{\v s}imir {\v S}ojat, Vanja {\v S}tefanec, Luigi Talamo, Delphine
  Tribout, Daniil Vodolazsky, Arseniy Vydrin, Aigul Zakirova, and Britta
  Zeller. 2022.
\newblock \href {http://hdl.handle.net/11234/1-4629} {Universal segmentations
  1.0 ({UniSegments} 1.0)}.
\newblock {LINDAT}/{CLARIAH}-{CZ} digital library at the Institute of Formal
  and Applied Linguistics ({{\'U}FAL}), Faculty of Mathematics and Physics,
  Charles University.

\bibitem[{Zeller et~al.(2013)Zeller, {\v{S}}najder, and
  Pad{\'o}}]{zeller2013derivbase}
Britta Zeller, Jan {\v{S}}najder, and Sebastian Pad{\'o}. 2013.
\newblock Derivbase: Inducing and evaluating a derivational morphology resource
  for german.
\newblock In \emph{Proceedings of the 51st Annual Meeting of the Association
  for Computational Linguistics (Volume 1: Long Papers)}, pages 1201--1211.

\bibitem[{Zeman et~al.(2017)Zeman, Popel, Straka, Hajic, Nivre, Ginter,
  Luotolahti, Pyysalo, Petrov, Potthast et~al.}]{zeman2017conll}
Daniel Zeman, Martin Popel, Milan Straka, Jan Hajic, Joakim Nivre, Filip
  Ginter, Juhani Luotolahti, Sampo Pyysalo, Slav Petrov, Martin Potthast,
  et~al. 2017.
\newblock Conll 2017 shared task: Multilingual parsing from raw text to
  universal dependencies.
\newblock In \emph{CoNLL 2017 Shared Task: Multilingual Parsing from Raw Text
  to Universal Dependencies}, pages 1--19. Association for Computational
  Linguistics.

\bibitem[{Zundui and Avaajargal(2022)}]{Task2_NUMDI}
Tsolmon Zundui and Chinbat Avaajargal. 2022.
\newblock Word-live morpheme segmentation using transformer neural network.
\newblock In \emph{19th SIGMORPHON Workshop on Computational Research in
  Phonetics, Phonology, and Morphology}.

\bibitem[{Žabokrtský et~al.(2022)Žabokrtský, Bafna, Bodnár, Kyjánek,
  Svoboda, Ševčíková, and Vidra}]{unisegments-lrec-2022}
Zdeněk Žabokrtský, Niyati Bafna, Jan Bodnár, Lukáš Kyjánek, Emil
  Svoboda, Magda Ševčíková, and Jonáš Vidra. 2022.
\newblock {Towards Universal Segmentations: UniSegments 1.0}.
\newblock In \emph{Proceedings of the 13th International Conference on Language
  Resources and Evaluation ({LREC} 2018)}, Marseille, France. European Language
  Resources Association (ELRA).

\end{thebibliography}
\bibliographystyle{acl_natbib}

\end{document}